\pdfoutput=1

\documentclass[11pt]{article}

\usepackage{acl}

\usepackage{times}
\usepackage{latexsym}

\usepackage[T1]{fontenc}

\usepackage[utf8]{inputenc}

\usepackage{microtype}

\usepackage{inconsolata}

\usepackage{graphicx}
\usepackage{caption}
\usepackage{subcaption}

\usepackage{algorithm}
\usepackage{algorithmic}

\usepackage{newfloat}
\usepackage{listings}
\DeclareCaptionStyle{ruled}{labelfont=normalfont,labelsep=colon,strut=off} 
\floatstyle{ruled}
\newfloat{listing}{tb}{lst}{}
\floatname{listing}{Algorithm}

\usepackage{booktabs}
\usepackage{multirow}
\usepackage{makecell}

\usepackage{adjustbox}

\usepackage{framed}

\usepackage{dsfont}
\usepackage{amsfonts}
\usepackage{bbm}
\usepackage{amsmath}
\usepackage{amsthm}
\usepackage{mathtools}

\DeclareMathOperator*{\argmax}{arg\,max}
\DeclareMathOperator*{\argmin}{arg\,min}

\usepackage[english]{babel}

\newtheorem{lemma}{Lemma}

\newcommand{\vx}{\mathbf{x}}
\newcommand{\vy}{\mathbf{y}}
\newcommand{\vh}{\mathbf{h}}

\newcommand{\Pmodel}{P_\mathrm{{model}}} 
\newcommand{\Pm}{P}

\newcommand{\CandH}{\mathcal{H}}
\newcommand{\RefH}{\mathcal{R}}

\usepackage{xcolor}

\title{
Hyperparameter-Free Approach for Faster Minimum Bayes Risk Decoding
}

\author{Yuu Jinnai \and Kaito Ariu\\ 
  CyberAgent \\
  \texttt{\{jinnai\_yu,kaito\_ariu\}@cyberagent.co.jp}
  }

\begin{document}
\maketitle
\begin{abstract}
Minimum Bayes-Risk (MBR) decoding is shown to be a powerful alternative to beam search decoding for a wide range of text generation tasks.
However, MBR requires a huge amount of time for inference to compute the MBR objective, which makes the method infeasible in many situations where response time is critical.
Confidence-based pruning (CBP) \cite{cheng-vlachos-2023-faster} has recently been proposed to reduce the inference time in machine translation tasks. Although it is shown to significantly reduce the amount of computation, it requires hyperparameter tuning using a development set to be effective. 
To this end, we propose Adaptive Minimum Bayes-Risk (AMBR) decoding, a hyperparameter-free method to run MBR decoding efficiently. 
AMBR is derived from the observation that the problem of computing the sample-based MBR objective is the \textit{medoid identification problem}.
AMBR uses the Correlated Sequential Halving (CSH) algorithm \cite{baharav2019ultra}, the algorithm with the best performance guarantee to date for the medoid identification problem, to compute the sample-based MBR objective.
We evaluate AMBR on machine translation, text summarization, and image captioning tasks. The results show that AMBR achieves on par with CBP, with CBP selecting hyperparameters through an Oracle for each given computation budget.
Our code is available at \url{https://github.com/CyberAgentAILab/adaptive-mbr}.
\end{abstract}

\section{Introduction}

The goal of natural language generation is to generate text representing structured information that is both fluent and contains the appropriate information. 
One of the key design decisions in text generation is the choice of decoding strategy. The decoding strategy is the decision rule used to generate sequences from a probabilistic language model.
Beam search has been widely used in many close-ended sequence generation tasks including machine translation \cite{wu2016googles,ott-etal-2019-fairseq,wolf-etal-2020-transformers}, text summarization \cite{rush-etal-2015-neural,narayan-etal-2018-dont}, and image captioning \cite{anderson-etal-2017-guided}.
However, beam search is known to have several degeneration problems. For example, \citet{welleck-etal-2020-consistency} reports that beam search can yield infinite-length outputs that the model assigns zero probability to.

\textbf{Minimum Bayes-Risk (MBR) decoding} has recently gained attention as a decoding strategy with the potential to overcome the problems of beam search \cite{goodman-1996-parsing,kumar-byrne-2004-minimum,eikema-aziz-2020-map,eikema-aziz-2022-sampling,freitag-etal-2022-high,bertschs2023}.
Unlike beam search which seeks to find the most probable output, MBR decoding seeks to find the output that maximizes the expected utility. MBR decoding involves two steps. It first samples outputs from the probabilistic model and then computes the utility between each pair of outputs to find the hypothesis with the highest expected utility.


One of the most important shortcomings of MBR decoding is its speed. 
The computational complexity of MBR decoding is $O(N \cdot G + N^2 \cdot U)$, where $N$ is the number of samples to be used, $G$ is the time to generate a sample, and $U$ is the time to evaluate the utility function. 
As the utility function is typically a time-consuming neural metric such as BLEURT and COMET \cite{sellam-etal-2020-bleurt,pu-etal-2021-learning,rei-etal-2020-unbabels,rei-etal-2022-comet}, $O(N^2 \cdot U)$ is the dominant factor of the computational complexity.



\textbf{Confidence-based pruning (CBP)} has recently proposed to reduce the number of evaluations of the utility function \cite{cheng-vlachos-2023-faster}. CBP is shown to be effective in machine translation tasks, significantly reducing the required computation using both lexical and neural utility functions with a negligible drop in the quality.

Although CBP is shown to be efficient, the performance of CBP is significantly influenced by the choice of hyperparameters. As such, CBP requires a development set for the tuning of these hyperparameters. Additionally, CBP cannot dictate the speed at which it completes tasks. The hyperparameters of CBP only offer indirect control over the number of evaluations. 

To this end, we propose \textbf{Adaptive Minimum Bayes-Risk (AMBR)} decoding, a hyperparameter-free algorithm to compute the sample-based MBR objective efficiently.
AMBR reformulates the MBR objective as the medoid identification problem \cite{rdusseeun1987clustering} and solves it using the Correlated Sequential Halving (CSH) algorithm, the best algorithm to date to solve the medoid identification problem \cite{baharav2019ultra}.
The strength of AMBR is that it is free from hyperparameters. 
Unlike CBP where it needs to tune the hyperparameters to empirically determine the best set of hyperparameters to achieve the desired trade-off between the speed and the quality, AMBR determines the best resource allocation automatically from the computational budget specified by the user.

We evaluate the performance of AMBR in machine translation, text summarization, and image captioning tasks.
The empirical results show that AMBR is on par with CBP with Oracle hyperparameters. They are roughly 4 to 8 times faster than MBR with a marginal drop in the output quality. 
The result indicates that using AMBR, MBR decoding can be run efficiently for a given computation budget specified on the fly without hyperparameter tuning on a development set.



\section{Background}
\label{sec:text}
Conditional text generation is the task of generating an output sequence $\vh$ given an input sequence $\vx$.
Probabilistic text generators define a probability distribution $\Pmodel (\vh | \vx)$ over an output space of hypotheses $\mathcal{Y}$.
In this paper, we denote $\Pmodel(\vh | \vx)$ by $\Pmodel(\vh)$ for brevity.
The goal of decoding is to find the highest-scoring hypothesis for a given input. 

One of the most common decision rules is maximum-a-posteriori (MAP) decoding. MAP decoding finds the most probable output under the model:
\begin{equation}
    \vh^{\mathrm{MAP}} = \argmax_{\vh \in \mathcal{Y}} \Pmodel(\vh).
\end{equation}
Although it seems intuitive to solve this MAP objective, prior work has pointed out two critical problems with this strategy. First, since the size of hypotheses set $|\mathcal{Y}|$ is extremely large, solving it exactly is intractable. Second, the MAP objective often leads to low-quality outputs \cite{stahlberg-byrne-2019-nmt,Holtzman2020The,meister-etal-2020-beam}. In fact, \citet{stahlberg-byrne-2019-nmt} shows that $\vh^{\mathrm{MAP}}$ is often the empty sequence in their experiment setting.

As such, beam search is commonly used as a heuristic algorithm to solve decoding problems \cite{graves2012sequence,Sutskever2014}. Beam search is known to generate higher-quality sequences than MAP decoding in a wide range of tasks.
Still, prior work has reported the degeneration issues of beam search such as repetitions and infinite-length outputs \cite{pmlr-v97-cohen19a,Holtzman2020The}.

\subsection{Minimum Bayes-Risk (MBR) Decoding}

Unlike MAP decoding which searches for the most probable output, MBR decoding seeks to find the output that maximizes the expected utility, thus minimizing the risk equivalently \cite{kumar-byrne-2002-minimum,kumar-byrne-2004-minimum}.
The procedure is made of two components: a machine translation model and a utility metric. The model $\Pmodel(\vy | \vx)$ estimates the probability of an output $\vy$ given an input sentence $\vx$. 
The utility metric $u(\vy, \vy')$ estimates the quality of a candidate translation $\vy$ given a reference translation $\vy'$.
Given a set of candidate hypotheses $\CandH \subseteq \mathcal{Y}$, we select the best hypothesis according to its expected utility with respect to the distribution of human references $P_\mathrm{human}$.
\begin{equation}
    \vh^{\mathrm{human}} = \argmax_{\vh \in \CandH}  \mathop{\mathbb{E}}_{\vy \sim P_\mathrm{{human}}} [u(\vh, \vy)].
\end{equation}
Because $P_\mathrm{human}$ is unknown, MBR instead uses the model probability $\Pmodel$ to approximate $P_\mathrm{human}$:
\begin{equation}
    \vh^{\mathrm{model}} = \argmax_{\vh \in \CandH} \mathop{\mathbb{E}}_{\vy \sim \Pmodel} [u(\vh, \vy)].
\label{eq:mbr}
\end{equation}
For the rest of the paper, we denote $\Pmodel$ as $\Pm$ for simplicity if not confusing.
As integration over $\mathcal{Y}$ is computationally intractable, Eq.~\eqref{eq:mbr} is approximated by a \textbf{Monte Carlo estimate} \cite{eikema-aziz-2022-sampling,farinhas2023empirical} using a pool of references $\RefH$ sampled from $\Pm$:
\begin{equation}
    \vh^{\mathrm{MC}} = \argmax_{\vh \in \CandH} \frac{1}{|\RefH|} \sum_{\vy \in \RefH} u(\vh, \vy).
\label{eq:empirical}
\end{equation}
In this paper, we investigate algorithms to compute $\vh^{\mathrm{MC}}$ efficiently. 

\subsection{Computational Complexity of MBR Decoding}

The shortcoming of the MBR is that it requires a huge amount of computation at inference time.
The computational complexity of MBR is $O(|\CandH \cup \RefH| \cdot G + |\CandH| |\RefH| \cdot U)$ where $G$ is the upper bound of the time to generate a hypothesis, and $U$ is the upper bound of the time to evaluate the utility function for a pair of hypotheses \cite{eikema-aziz-2022-sampling}.
Sample-based MBR typically uses the same set of hypotheses for the candidate set $\CandH$ and the reference pool $\RefH$ ($\mathcal{H} = \RefH$). In this way the computational complexity is $O(N \cdot G + N^2 \cdot U)$, where $N = |\mathcal{H}| = |\mathcal{R}|$.
Thus, The bottleneck of the computation is typically the evaluation of the utility function.

Several approaches have been proposed to improve the efficiency of MBR decoding before confidence-based pruning \cite{eikema-aziz-2022-sampling,freitag-etal-2022-high}. \textbf{N-by-S (NbyS)} seeks to reduce the total number of evaluations by reducing the reference pool \cite{eikema-aziz-2022-sampling}. 
\citet{eikema-aziz-2022-sampling} provides empirical evidence showing that increasing the number of candidates is more effective than increasing the number of references.
The computational complexity of N-by-S with $S' (< N)$ references is $O(N \cdot G + N S' \cdot U)$.
\textbf{Coarse-to-Fine (C2F)} reduces the size of the candidate and reference hypotheses using a coarse utility function \cite{eikema-aziz-2022-sampling}. It first runs coarse evaluation using a faster utility function (e.g. non-neural lexical scoring function). It then selects the top-scoring hypotheses as a pruned candidate set and reference set. Finally, it runs the MBR decoding with the finer utility function using the pruned candidate and reference set to output the best hypothesis.
In this way, the total computation required by C2F is $O(N \cdot G + N^2 \cdot U' + N' S' \cdot U)$ where $U'$ is the computational cost of the coarse utility function, $N', S' (\leq N)$ are the size of the pruned candidate and reference set.

\textbf{Reference Aggregation (RA)} computes the MBR score against aggregated reference representations to reduce the computational complexity to $O(N \cdot G + N \cdot U^A)$, where $U^A$ is the upper bound on the complexity of evaluating the aggregated utility function \cite{vamvas2024lineartime}. 
The shortcoming of RA is that it is not applicable to non-aggregatable utility functions.
For example, MetricX-23 \cite{juraska-etal-2023-metricx} is a transformer-based metric where the input is a sequence of embeddings of the tokens instead of the embedding of the whole sentence, making it non-aggregatable.
Another example is where the utility function involves a reward function. See Appendix \ref{sec:reward} for details.

\section{Confidence-Based Pruning (CBP)}
\textbf{Confidence-based pruning (CBP)} is recently proposed by \citet{cheng-vlachos-2023-faster} to significantly reduce the number of evaluations of the utility function.
The idea is to iteratively evaluate the hypotheses with a subset of the reference set to prune the hypotheses not promising enough. 

CBP keeps a current candidate set $\CandH_i$ and a current reference set $\RefH_i$ during the run. The candidate set starts from the whole candidates ($\CandH_0 = \mathcal{H}$) and the reference set starts empty ($\RefH_0 = \emptyset$). 
At every iteration $i$, it draws samples and adds them to the reference set until the size of the reference set reaches the limit $r_i$, where $\{r_i\}$ are hyperparameters.
Then it computes the incumbent best solution $\vh_i^*$ at $i$-th iteration:
\begin{equation}
    \vh_i^* = \argmax_{\vh \in \CandH_i} \frac{1}{|\RefH_i|} \sum_{\vy \in \RefH_{i}} u(\vh, \vy).
\end{equation}
Then it generates a series of bootstrap reference sets $\hat{\RefH}_i^b$ which is a with-replacement size-$|\RefH|$ resample of $\RefH_i$. Using a series of bootstrap reference sets, it computes the estimated win ratio of each hypothesis against $\vh_i^*$ in $\CandH_i$:
\begin{equation}
    w(\vh) = \frac{1}{B}\sum_{b=1}^{B} \mathbbm{1}[\sum_{\vy \in \hat{\RefH}_i^b} u(\vh, \vy) \geq \sum_{\vy \in \hat{\RefH}_i^b} u(\vh_i^*, \vy)],
\end{equation}
where $B$ is the number of bootstrap reference sets.
Then, it prunes all candidates from the candidate set with the win ratio lower than $1-\alpha$, where $\alpha$ is a hyperparameter. It repeats the process until the size of the candidate set reaches $1$ or the sample size scheduler terminates.

Although CBP is shown to be significantly more efficient than the standard MBR, there are several shortcomings.
First, it requires a hyperparameter tuning using the development set. The sample size scheduler $r_i$ and the confidence threshold $\alpha$ need to be tuned to optimize the performance. 
The number of bootstrap reference sets $B$ is also a hyperparameter that needs to be tuned according to the quality and the speed trade-off.
Note that the optimal set of hyperparameters is influenced by the desired speed-up. If one wants to choose 2x speed-up and 4x speed-up according to the situation, one needs to search for two sets of hyperparameters for each budget constraint.
Additionally, CBP cannot give a budget constraint and optimize under that. Because the hyperparameters of CBP only indirectly control the number of evaluations it needs to finish, a user has no direct control over the desired speed-up.

\section{Adaptive Minimum Bayes Risk (AMBR) Decoding}


\begin{listing}
\caption{Adaptive MBR (AMBR) \\ (Correlated Sequential Halving for MBR)}
\label{lst:ambr}
\renewcommand{\algorithmicrequire}{\textbf{Input:}}
\renewcommand{\algorithmicensure}{\textbf{Output:}}
\begin{algorithmic}[1]
\REQUIRE a set of candidates $\mathcal{H}$, references $\mathcal{R}$, and a budget $T$ \\
\ENSURE a hypothesis $\vh^{\mathrm{AMBR}}$ \\
\STATE $\CandH_0 \leftarrow \mathcal{H}$ \\
\STATE $\RefH_0 \leftarrow \emptyset$ \\
\STATE $N = \max(|\mathcal{H}|, |\mathcal{R}|)$ \\
\FOR {$i = 0$ \textbf{to} $\lceil \log N \rceil - 1$}
    \STATE $t_i = \min(\max(\lfloor \frac{T}{|\CandH_i| \lceil \log n \rceil} \rfloor, 1), n)$ \\
    \STATE Let $J_i$ be a set of $t_i - |\RefH_i|$ references sampled from $\mathcal{R} \setminus \RefH_i$ without replacement \\
    \STATE $\RefH_{i+1} = J_i \cup \RefH_i$ \\
    \FOR {$\vh \in \CandH_i$}
        \STATE $\hat{U}(\vh) \leftarrow \frac{1}{|\RefH_{i+1}|} \sum_{\vy \in \RefH_{i+1}} u(\vh, \vy)$ \\
    \ENDFOR
    \IF {$t_i = n$}
        \RETURN $\argmax_{\vh \in \CandH_i} \hat{U}(\vh)$ \\
    \ELSE
        \STATE Let $\CandH_{i+1}$ be the set of $\lceil |\CandH_i|/2 \rceil$ candidates in $\CandH_i$ with the largest $\hat{U}(\vh)$
    \ENDIF
\ENDFOR
\RETURN $\argmax_{\vh \in \CandH^i} \hat{U}(\vh)$
\end{algorithmic}
\end{listing}

We propose \textbf{Adaptive Minimum Bayes-Risk (AMBR)} decoding, a variant of MBR that can efficiently compute the MBR objective under a budget on the maximum number of evaluations that a user can specify. The advantages of AMBR over CBP are twofold.
First, AMBR has no hyperparameter. The schedules of the number of references and the candidates are automatically determined by the algorithm. 
Second, a user can enforce the upper bound of the computation budget to AMBR. 
AMBR enforces the budget constraint and the algorithm automatically schedules how to use the limited resource accordingly.

AMBR is derived from the observation that MBR decoding is the \textit{medoid identification problem} \cite{pam1990}: the problem of computing $\vh^{\mathrm{MC}}$ (Eq~\ref{eq:empirical}) is tantamount to determining the medoid of $\mathcal{H}$.
The medoid, denoted as $\vy^*$, is defined as the point in a dataset $Y$ that minimizes the sum of distances to all other points:\footnote{The formulation of Eq.~\eqref{eq:medoid} represents the same class of problem as the standard formulation of medoid identification problem where it assumes $X = Y$. See Appendix \ref{sec:medoid} for the details.}
\begin{equation}
    \vx = \argmin_{\vx \in X} \sum_{\vy \in Y} d(\vx, \vy).
\label{eq:medoid}
\end{equation}
Let $d = -u$, $X = \mathcal{H}$, and $Y = \mathcal{R}$. Then, the problem can be translated into the following:
\begin{equation}
    \vy^* = \argmax_{\vy \in \mathcal{H}} \sum_{\vy' \in \mathcal{R}} u(\vy, \vy').
\end{equation}
This is exactly the objective defined in Eq.~\eqref{eq:empirical}. 

Our approach is to use the best algorithm proposed so far for solving the medoid identification problem and repurpose it for MBR decoding.
The algorithm with the best performance guarantee to date for solving the medoid identification is the \textbf{Correlated Sequential Halving (CSH)} algorithm \cite{baharav2019ultra}. 
We describe the procedure of AMBR in Algorithm~\ref{lst:ambr}. 
AMBR keeps a current candidate set $\CandH_i$ which starts with $\CandH$ and a current reference set $\RefH_i$ which starts as an empty set.
First, it picks $t_i$ hypotheses from $\RefH$ and adds them to the current reference set $\RefH_{i+1}$ where $t_i$ is automatically determined by the number of candidates and the budget. Then it computes $u(\vh, \vy)$ for all $\vh$ in the current candidate set $\CandH_i$ and for all $\vy$ in the current reference set $\RefH_{i+1}$. The average utility of $\vh \in \CandH_i$ over the current reference set is stored in $\hat{U}(\vh)$. Then, it runs the halving operation, pruning the lower half of the candidates according to the current estimate $\hat{U}$. Ties are broken arbitrarily. It repeats this process for up to $\lceil \log N \rceil - 1$ times and returns the candidate with the best estimate in $\CandH_i$ at that point.

The procedure of Algorithm \ref{lst:ambr} is identical to the procedure of CSH with modification to the notations to place it in the context of the decoding problem.
Our contribution is the reinvention of CSH which is proposed as a solution to the medoid identification problem as a tool to compute the MBR objective adaptively by converting the sum of distances to the expected utility.

\subsection{Analytical Result}

CSH has a theoretical guarantee of the probability of choosing the hypothesis with the highest utility in its original form \cite{baharav2019ultra}. 
The original form of CSH is recovered by replacing Line 7 of Algorithm~\ref{lst:ambr} with the following equation:
\begin{equation}
    \RefH_{i+1} = J_i.
\label{eq:guaranteed}
\end{equation}
AMBR using Eq~\eqref{eq:guaranteed} (AMBR-Replace) inherits the theoretical guarantee of CSH:
\begin{lemma}
Assuming $T \geq N \log N$, AMBR replacing Line 7 with Eq.~\eqref{eq:guaranteed} (AMBR-Replace) correctly identifies $\vh^{\mathrm{MC}}$ with probability at least $1 - \log N \exp(- \frac{T}{\log N} C)$ where $C$ is an instance dependent variable determined by $u$ and $\mathcal{H}$.
\label{lm:ambr}
\end{lemma}
See Theorem 2.1. of \citet{baharav2019ultra} for proof and a detailed description of the instance-dependent variable $C$.

Note that the theoretical guarantee is proven for AMBR-Replace (AMBR using Eq.~\ref{eq:guaranteed}). However, it has not been proven for Algorithm~\ref{lst:ambr} (AMBR).
Still, we remain optimistic that a similar guarantee holds for AMBR. To the best of our knowledge, no counterexamples exist to disprove this. The theoretical guarantee is thus proven to AMBR-Replace which is an inferior version of AMBR. The inferiority of AMBR-Replace stems from its process of discarding samples at every iteration, leading us to expect a less accurate estimate of the candidates' value. Therefore, we anticipate that the proposed algorithm outperforms the algorithm with a theoretical guarantee. In this way, although we lack a formal guarantee for AMBR, we do have a guarantee for a less effective version of it. This provides an informal justification for the effectiveness of AMBR.
This derivation is noted in Remark~1 in \citet{baharav2019ultra}.




\begin{figure*}[htb]
     \centering
     \begin{subfigure}[b]{0.41\textwidth}
         \centering
         \includegraphics[width=\textwidth]{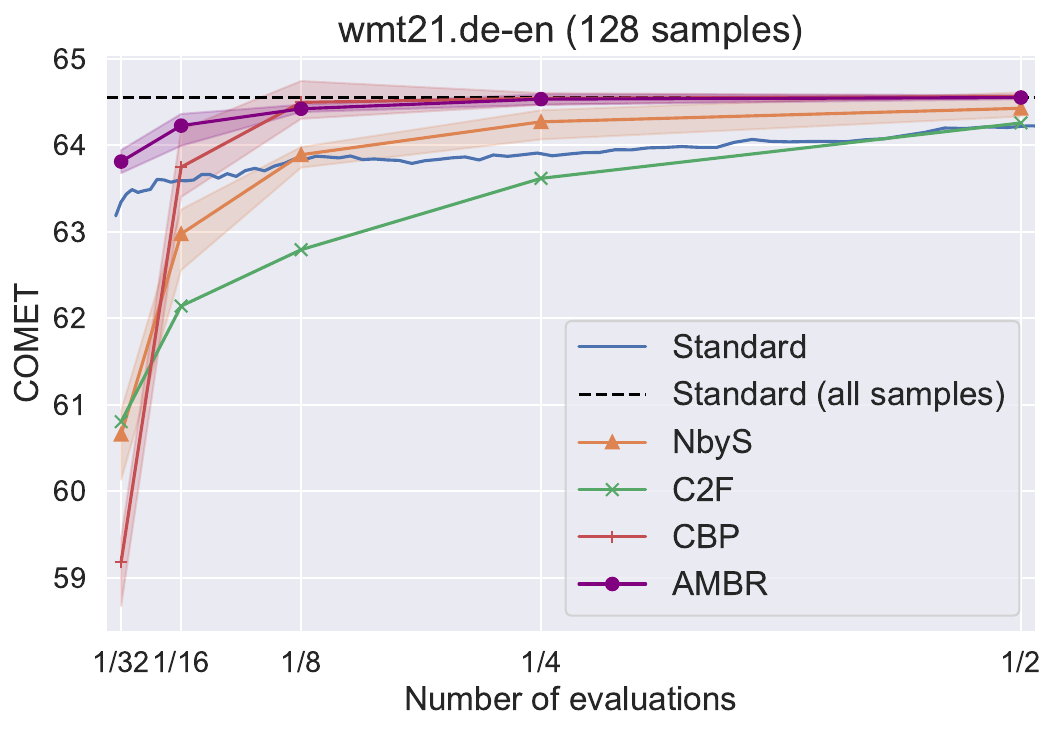}
         \caption{COMET-20 $\uparrow$ (De-En, $N=128$)}
         \label{fig:deen}
     \end{subfigure}
     \hspace{24pt}
     \begin{subfigure}[b]{0.41\textwidth}
         \centering
         \includegraphics[width=\textwidth]{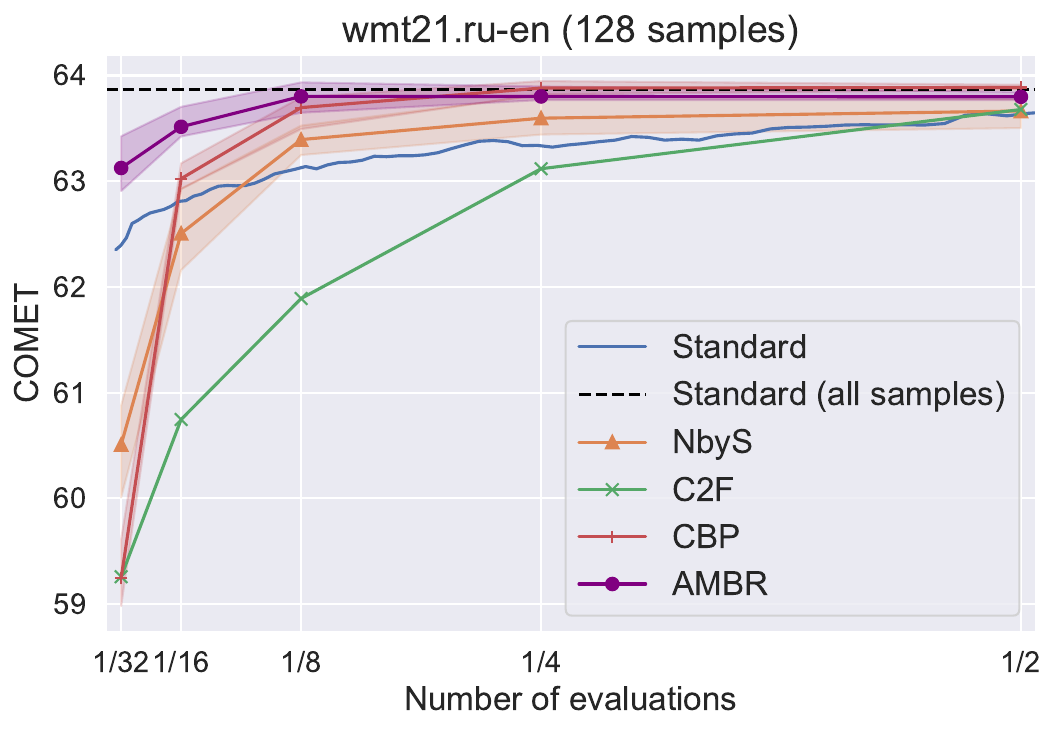}
         \caption{COMET-20 $\uparrow$ (Ru-En, $N=128$)}
         \label{fig:ruen}
     \end{subfigure}\\
     \begin{subfigure}[b]{0.41\textwidth}
         \centering
         \includegraphics[width=\textwidth]{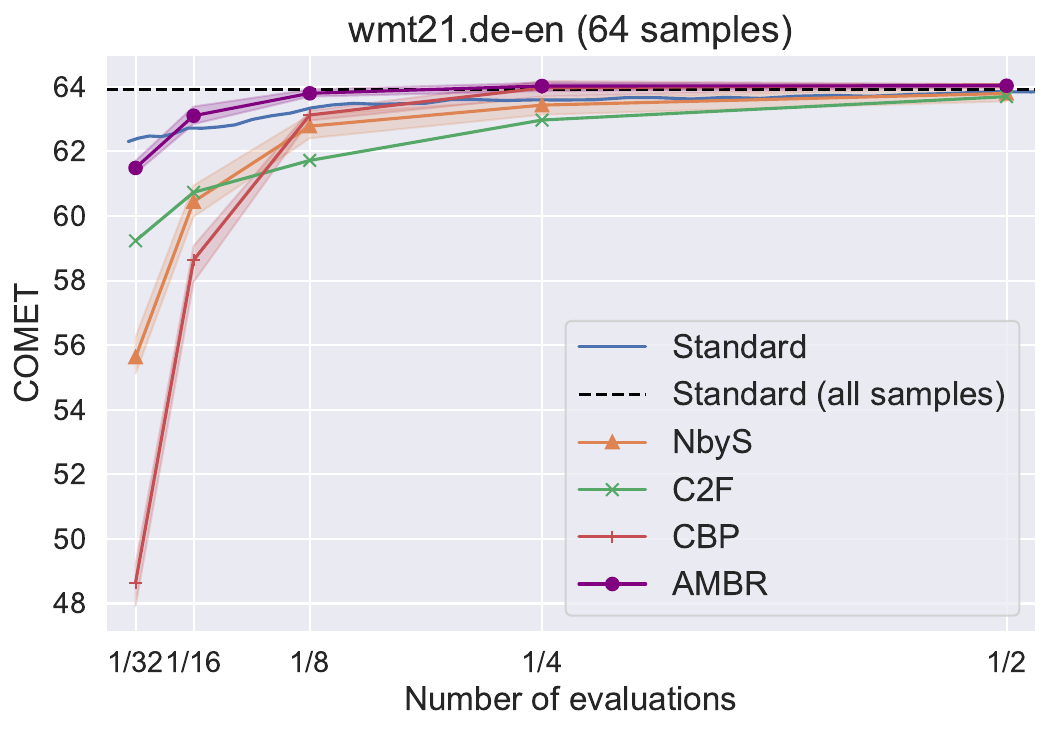}
         \caption{COMET-20 $\uparrow$ (De-En, $N=64$)}
         \label{fig:deen64}
     \end{subfigure}
     \hspace{24pt}
     \begin{subfigure}[b]{0.41\textwidth}
         \centering
         \includegraphics[width=\textwidth]{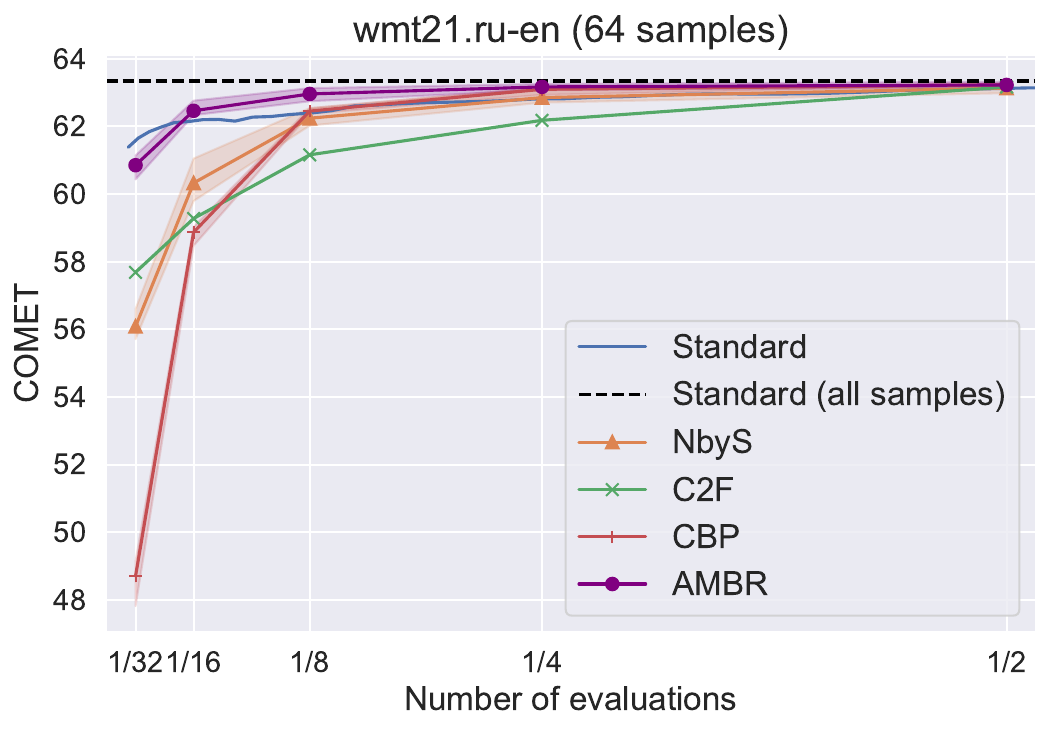}
         \caption{COMET-20 $\uparrow$ (Ru-En, $N=64$)}
         \label{fig:ruen64}
     \end{subfigure}
    \caption{COMET-20 score on WMT'21 De-En and Ru-En using the WMT 21 X-En model. The shaded regions show the minimum and the maximum values over five runs. The horizontal axis shows the reduction in the number of evaluations compared to the standard MBR with all samples.}
    \label{fig:mt}
\end{figure*}


\section{Experiments}
\label{sec:experiments}

\begin{figure*}
     \centering
     \begin{subfigure}[b]{0.32\textwidth}
         \centering
         \includegraphics[width=\textwidth]{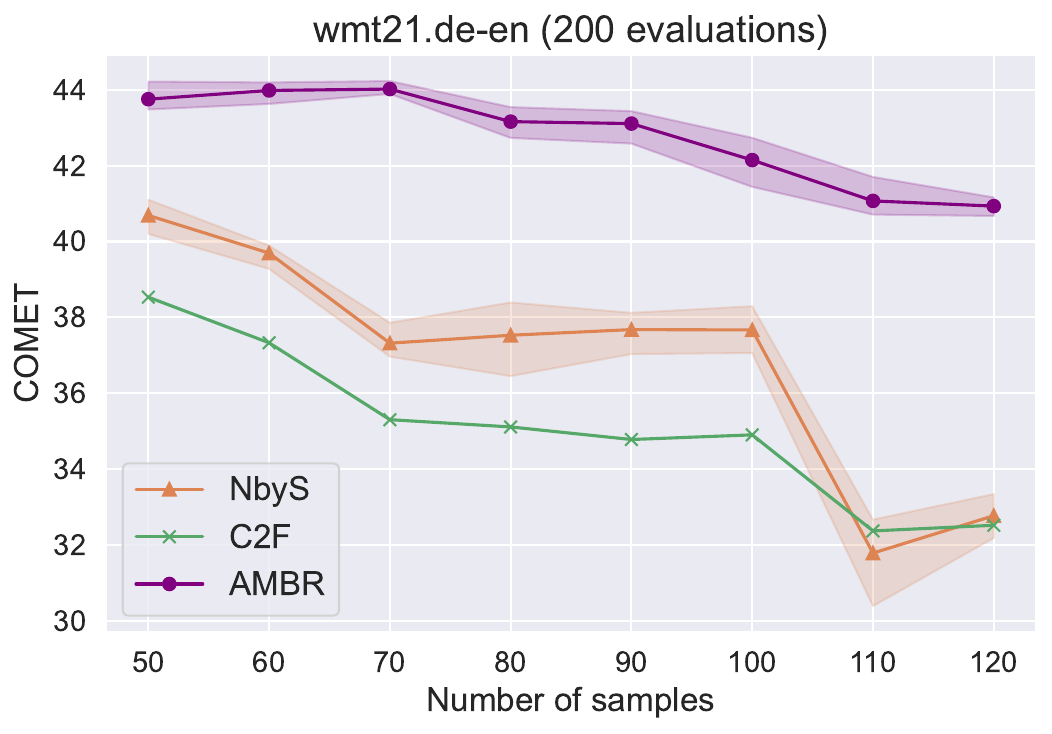}
         \caption{COMET-20 $\uparrow$ (De-En, $200$ evaluations)}
         \label{fig:deen200}
     \end{subfigure}
     \hfill
     \begin{subfigure}[b]{0.32\textwidth}
         \centering
         \includegraphics[width=\textwidth]{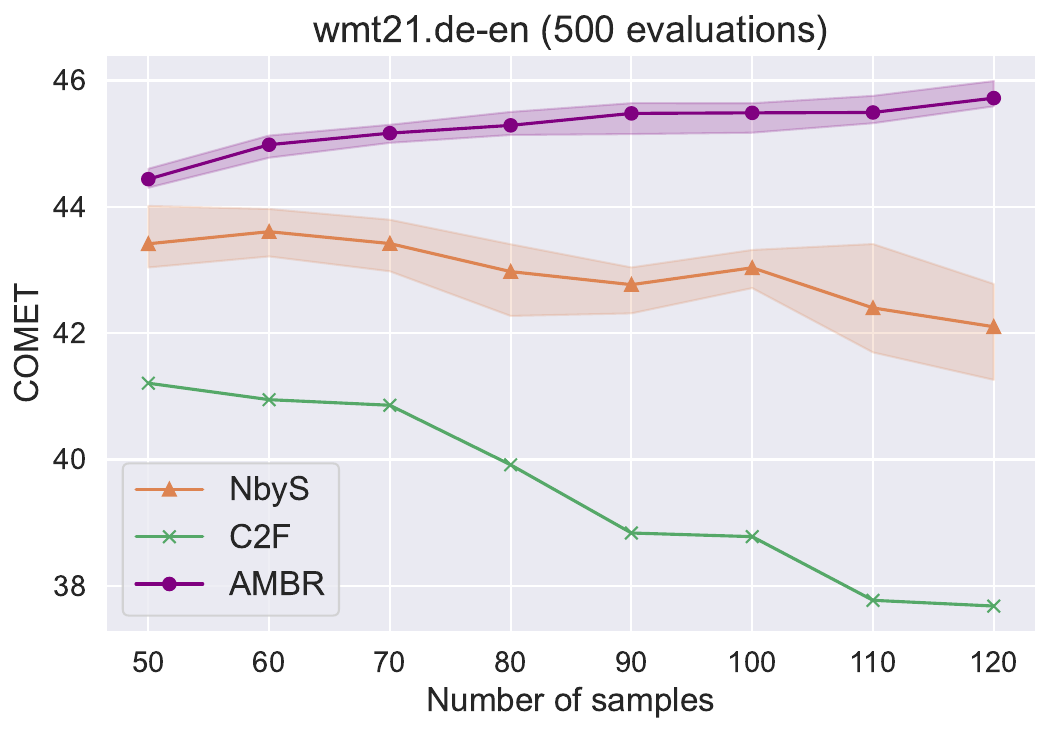}
         \caption{COMET-20 $\uparrow$ (De-En, $500$ evaluations)}
         \label{fig:deen500}
     \end{subfigure}
     \hfill
     \begin{subfigure}[b]{0.32\textwidth}
         \centering
         \includegraphics[width=\textwidth]{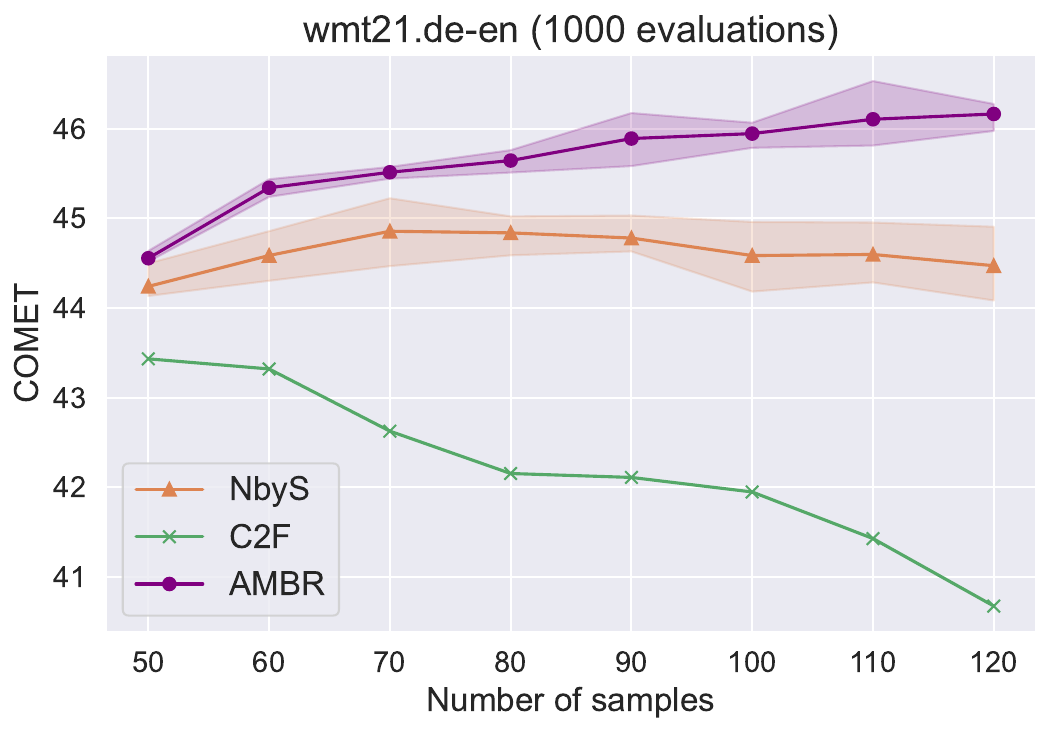}
         \caption{COMET-20 $\uparrow$ (De-En, $1000$ evaluations)}
         \label{fig:deen1000}
     \end{subfigure}
     
     \begin{subfigure}[c]{0.32\textwidth}
         \centering
         \includegraphics[width=\textwidth]{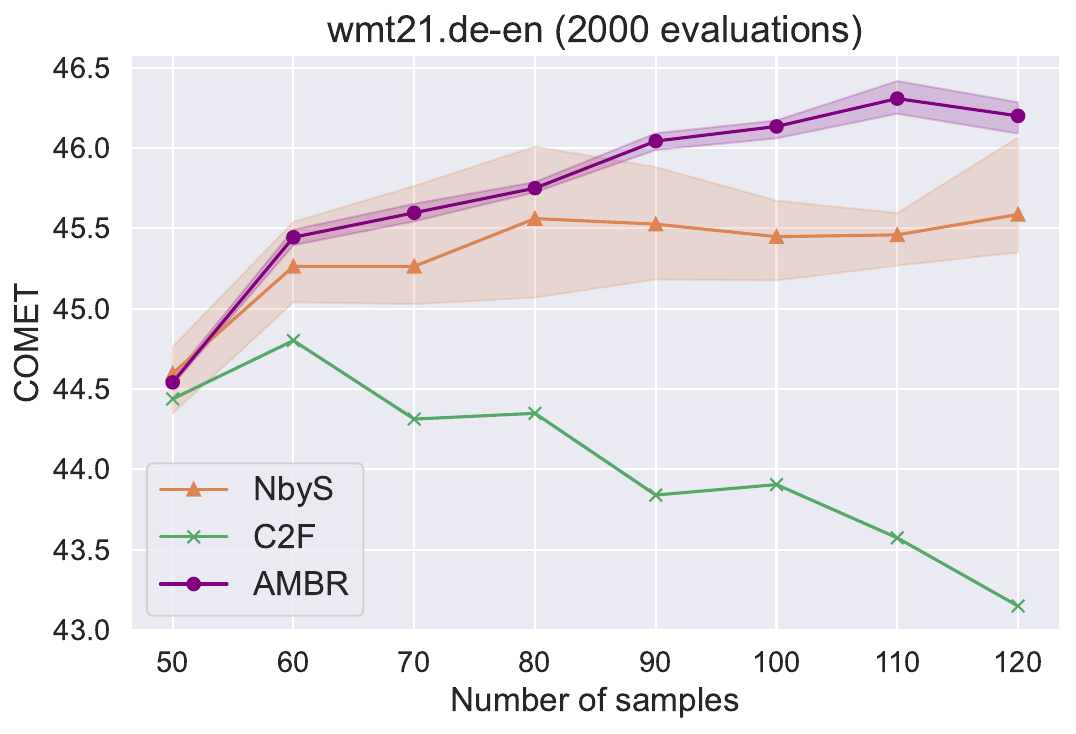}
         \caption{COMET-20 $\uparrow$ (De-En, $2000$ evaluations)}
         \label{fig:deen2000}
     \end{subfigure}
     \hspace{10pt}
     \begin{subfigure}[c]{0.32\textwidth}
         \centering
         \includegraphics[width=\textwidth]{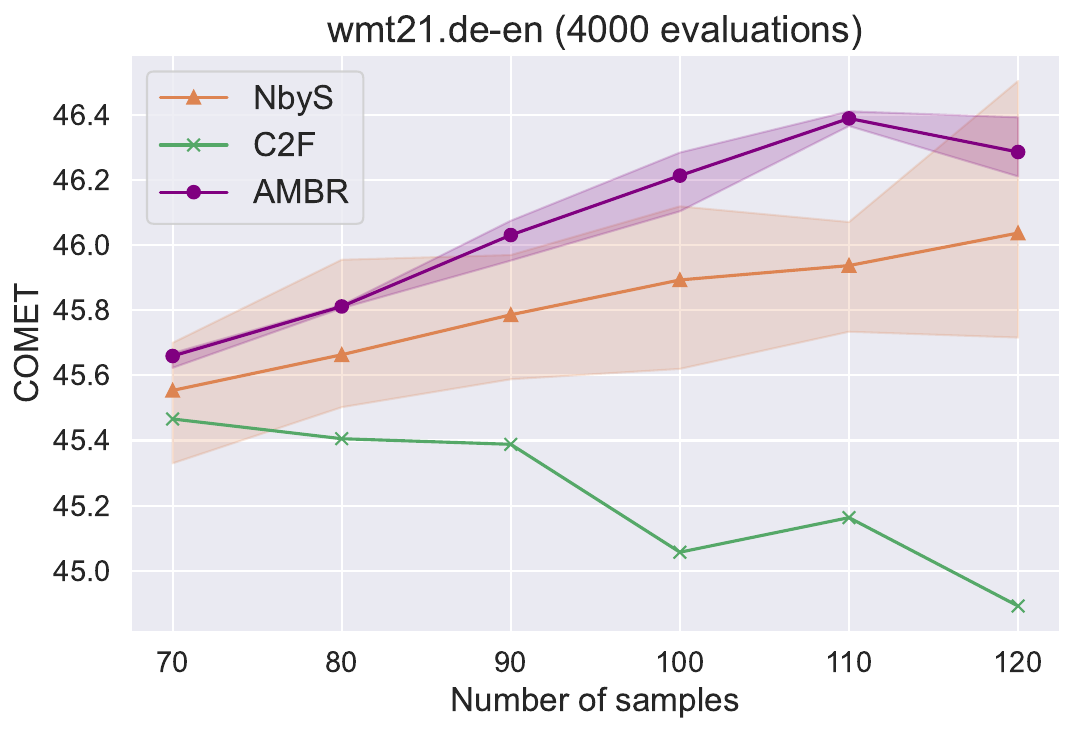}
         \caption{COMET-20 $\uparrow$ (De-En, $4000$ evaluations)}
         \label{fig:deen4000}
     \end{subfigure}
    \caption{Evaluation of AMBR with a varying number of samples with a fixed evaluation budget on WMT'21 De-En with COMET-20 score using the M2M100 418M model. The shaded regions show the minimum and the maximum values over five runs.}
    \label{fig:samples}
\end{figure*}

We evaluate the performance of the efficient MBR decoding algorithms on machine translation, text summarization, and image captioning tasks.
We evaluate the performance of the MBR decoding algorithms under a budget constraint on the number of evaluations. 
We evaluate with a budget size of $1/32, 1/16, 1/8, 1/4, 1/2$ of $N(N-1)$, the number of evaluations of the standard $N \times N$ MBR with $N$ samples.\footnote{We assume $u(\vh, \vh)$ is a constant for all $\vh \in \mathcal{H}$.}
We use epsilon sampling with $\epsilon=0.02$ as a sampling algorithm \cite{hewitt-etal-2022-truncation,freitag2023epsilon}. Temperature is fixed to $1.0$. We use the same set of samples for all the algorithms. 

We compare the performance of (Standard) MBR, N-by-S (NbyS), Coarse-to-fine (C2F), confidence-based pruning (CBP), and AMBR. 
Standard MBR refers to the implementation of MBR which uses the same set of samples for the candidate and reference set. We run standard MBR with the number of samples $N' \in \{1...N\}$. The number of evaluations for standard MBR is $N' (N' - 1)$.
We implement N-by-S in a way that uses all the samples $\mathcal{H}$ as the candidate set and reduces the size of references according to the budget. That is, it randomly subsamples $S'$ hypotheses from $\mathcal{H}$ to be the reference set so that $S'$ is the smallest integer such that $(N - 1) S' \geq T$. 
For C2F, we set $S' = N$ and $N'$ to be the smallest integer such that $N' (N - 1) \geq T$.
We run a hyperparameter sweep for CBP to find the best hyperparameters. We search over $r_0 \in \{1, 2, 4, 8\}$ and $\alpha \in \{0.8, 0.9, 0.99\}$. Following \citet{cheng-vlachos-2023-faster}, we set the schedule of the size of the references $r_i$ to double each step: $r_i = 2^i r_0$. The number of bootstrap reference sets is $500$.
We enforce the budget constraint to CBP by terminating the iteration once the number of evaluations reaches $T$. We run CBP with each set of hyperparameters on the test set to find the best hyperparameters. The result of the hyperparameter search is described in Appendix \ref{sec:cbp-params}. We observe that the best set of hyperparameters of CBP is dependent on the size of the budget. As such, we report the Oracle score, the best score over all combinations of hyperparameters for each budget.
AMBR is implemented as in Algorithm~\ref{lst:ambr} without using Eq.~\eqref{eq:guaranteed}. Thus, Lemma~\ref{lm:ambr} does not apply to the algorithm we evaluate in this section.
We run NbyS, CBP, and AMBR five times for each budget size and report the average, minimum, and maximum scores over the runs.

We use Huggingface's Transformers library for running all the experiments \cite{wolf-etal-2020-transformers}.
All the experiments are conducted using publicly available pretrained models and datasets for reproducibility.
Due to limitations in computational resources, we evaluate the first 1000 entries of each dataset.

\begin{figure*}[!tbh]
    \centering
     \begin{subfigure}[b]{0.325\textwidth}
         \centering
         \includegraphics[width=\textwidth]{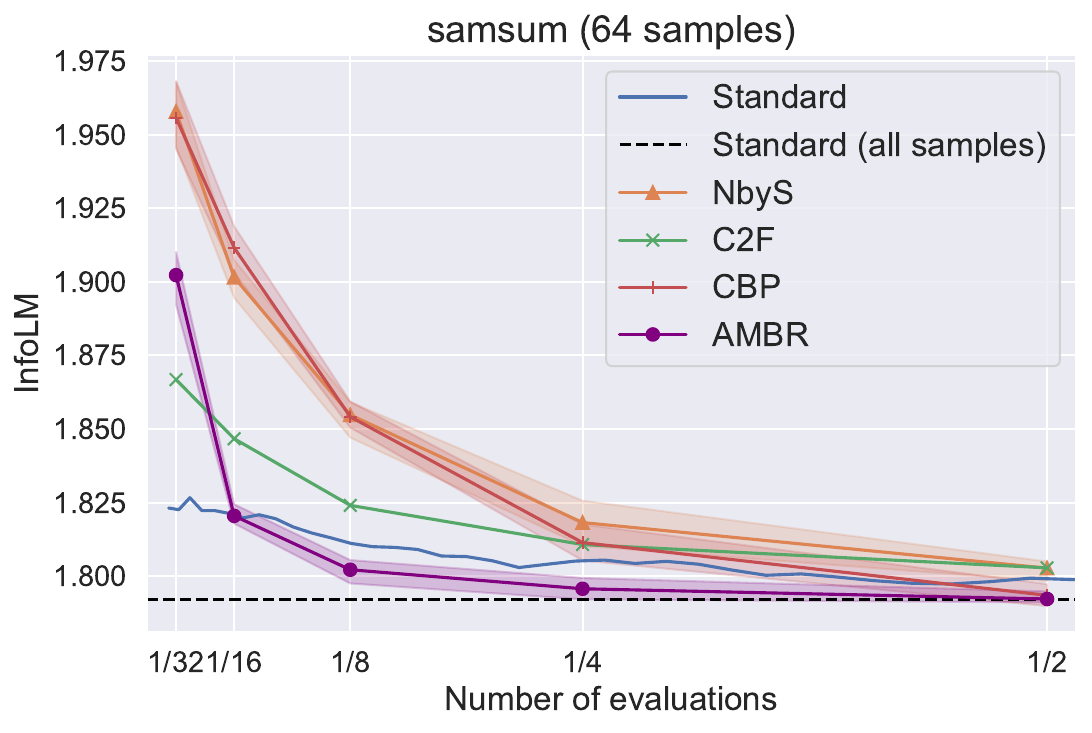}
         \caption{InfoLM $\downarrow$ (SAMSum)}
         \label{fig:samsum}
     \end{subfigure}
     \hfill
     \begin{subfigure}[b]{0.325\textwidth}
         \centering
         \includegraphics[width=\textwidth]{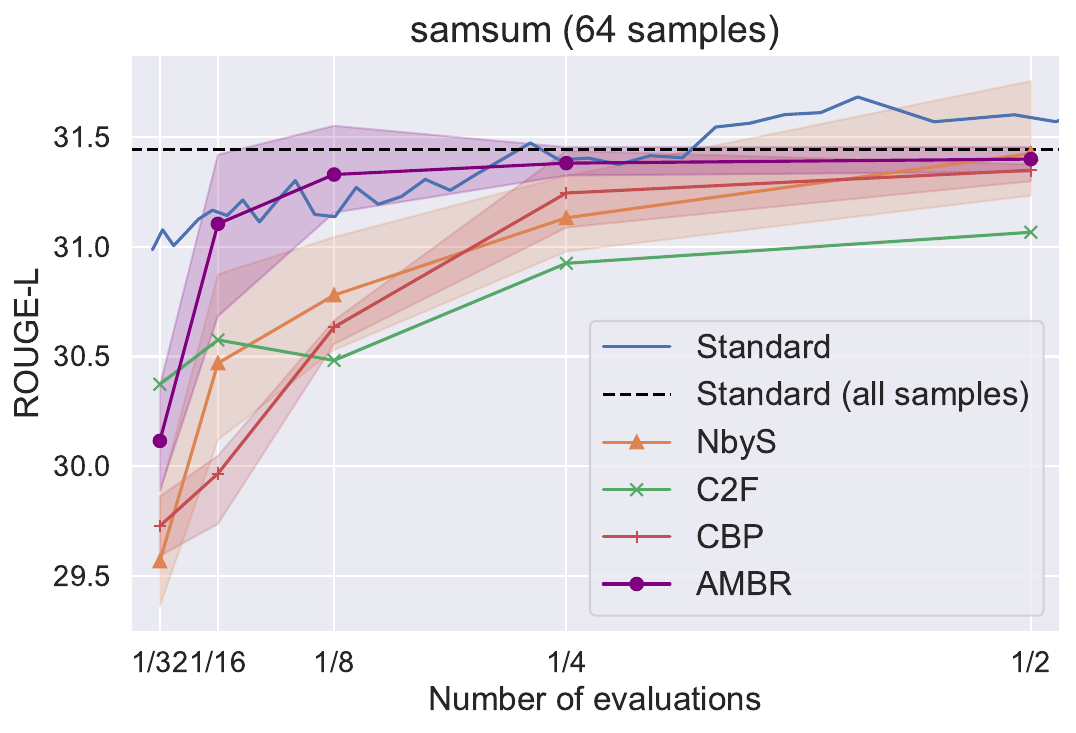}
         \caption{ROUGE-L $\uparrow$ (SAMSum)}
         \label{fig:samsum-rouge}
     \end{subfigure}
     \hfill
      \begin{subfigure}[b]{0.325\textwidth}
         \centering
         \includegraphics[width=\textwidth]{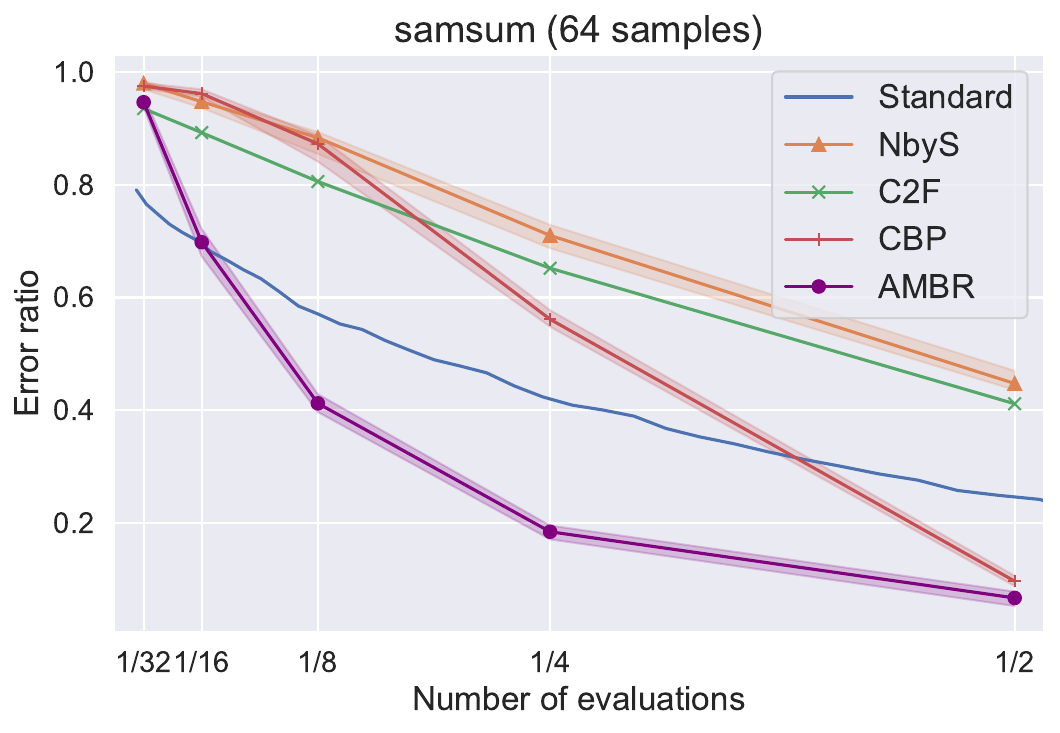}
         \caption{Error Rate $\downarrow$ (SAMSum)}
         \label{fig:samsum-err}
     \end{subfigure} \\
     \begin{subfigure}[b]{0.325\textwidth}
         \centering
         \includegraphics[width=\textwidth]{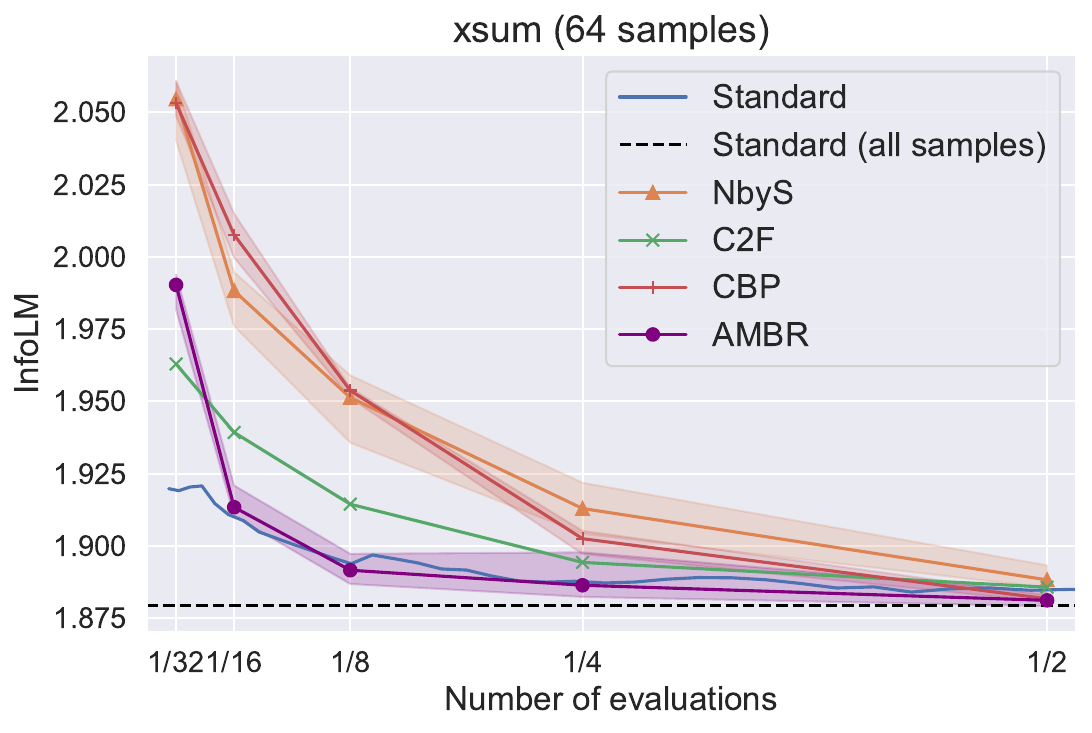}
         \caption{InfoLM $\downarrow$ (XSum)}
         \label{fig:xsum}
     \end{subfigure}
     \hfill
     \begin{subfigure}[b]{0.325\textwidth}
         \centering
         \includegraphics[width=\textwidth]{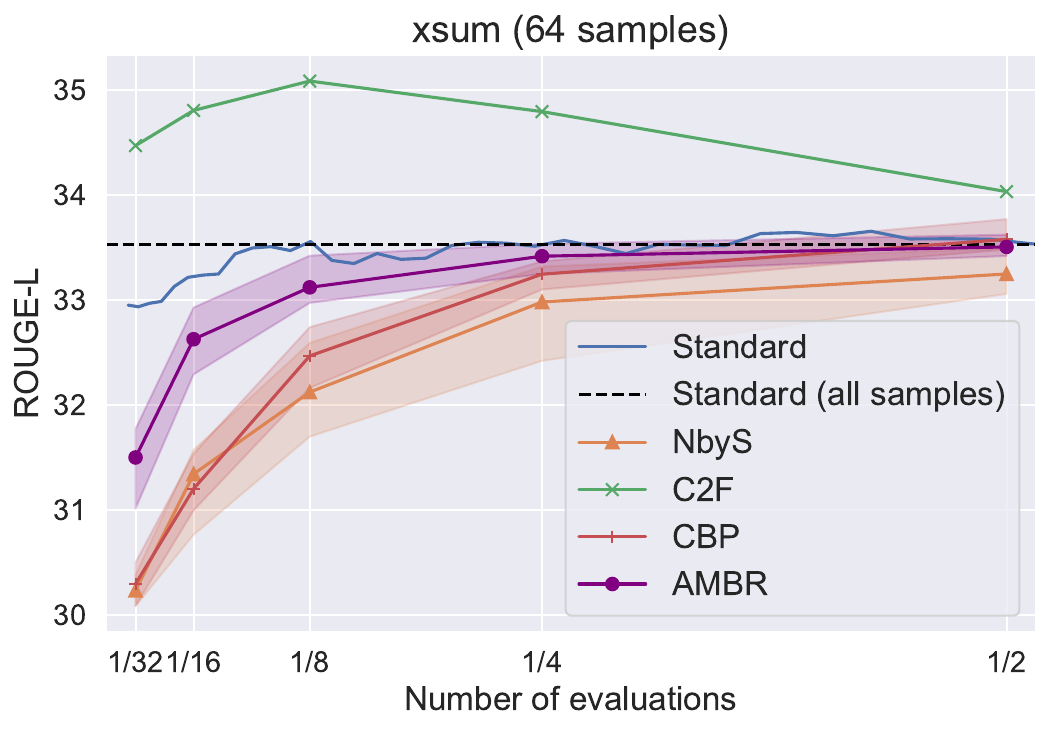}
         \caption{ROUGE-L $\uparrow$ (XSum)}
         \label{fig:xsum-rouge}
     \end{subfigure}
     \hfill
     \begin{subfigure}[b]{0.325\textwidth}
         \centering
         \includegraphics[width=\textwidth]{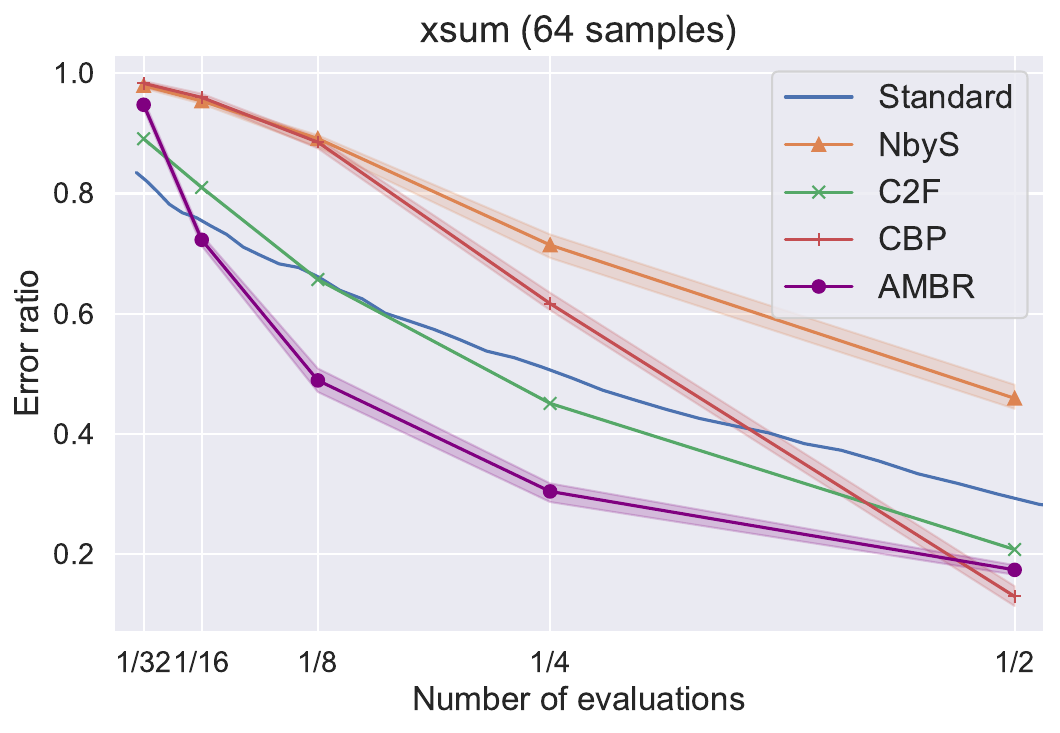}
         \caption{Error Rate $\downarrow$ (XSum)}
         \label{fig:xsum-err}
     \end{subfigure} \\
    \begin{subfigure}[b]{0.325\textwidth}
         \centering
         \includegraphics[width=\textwidth]{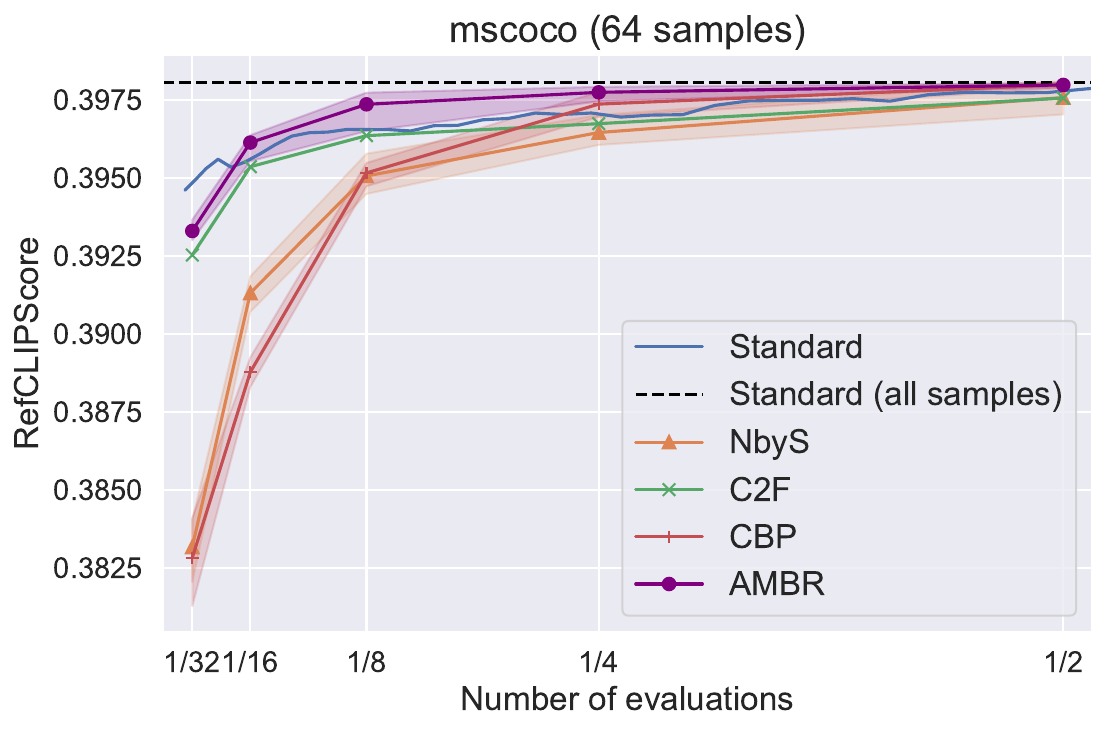}
         \caption{RefCLIPScore $\uparrow$ (MS COCO)}
         \label{fig:mscoco}
     \end{subfigure}
     \hfill
     \begin{subfigure}[b]{0.325\textwidth}
         \centering
         \includegraphics[width=\textwidth]{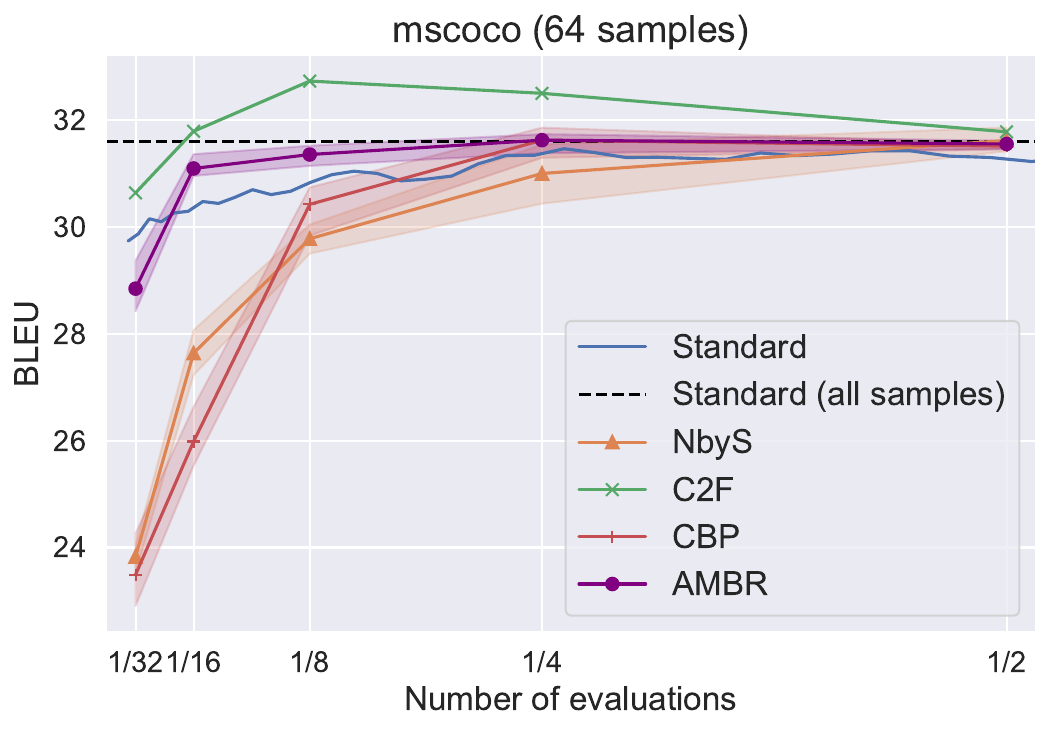}
         \caption{BLEU $\uparrow$ (MS COCO)}
         \label{fig:mscoco-bleu}
     \end{subfigure}
     \hfill
     \begin{subfigure}[b]{0.325\textwidth}
         \centering
         \includegraphics[width=\textwidth]{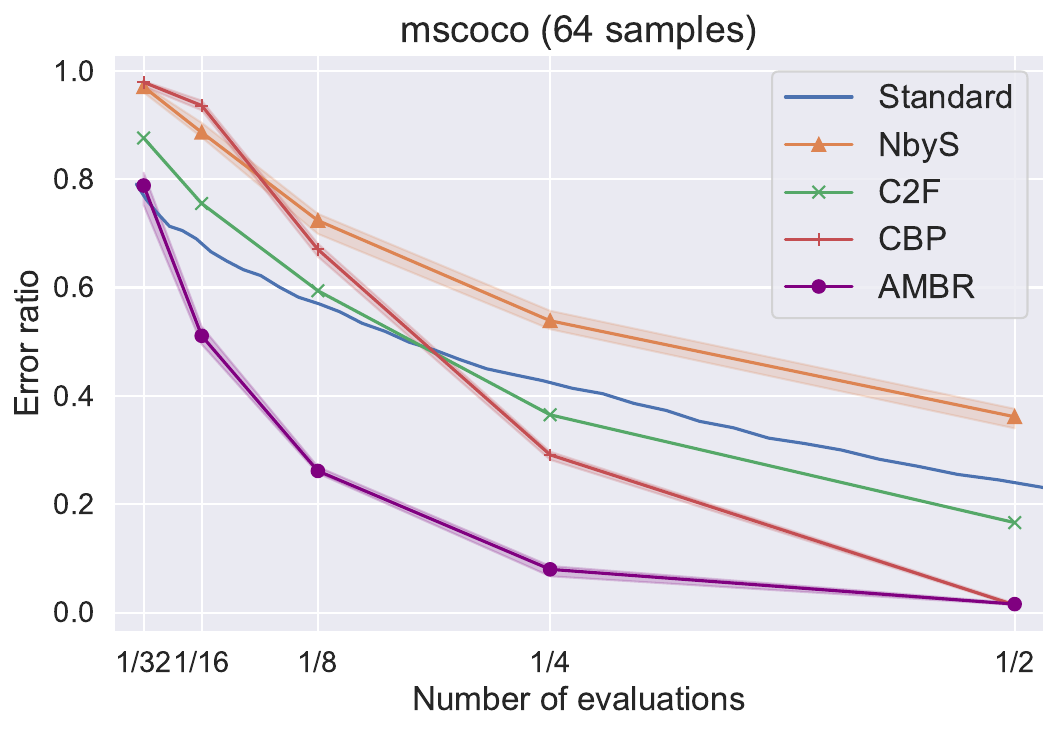}
         \caption{Error Rate $\downarrow$ (MS COCO)}
         \label{fig:mscoco-err}
     \end{subfigure}
    \caption{(\subref{fig:samsum}) InfoLM score, (\subref{fig:samsum-rouge}) ROUGE-L score, and (\subref{fig:samsum-err}) error rate on SAMSum dataset.
    (\subref{fig:xsum}) InfoLM score, (\subref{fig:xsum-rouge}) ROUGE-L score, and (\subref{fig:xsum-err}) error rate on XSum dataset.
    (\subref{fig:mscoco}) RefCLIPScore, (\subref{fig:mscoco-bleu}) BLEU score, and (\subref{fig:mscoco-err}) error rate on MS COCO dataset.
    The shaded regions show the minimum and the maximum values over five runs. The error rate is defined as the ratio of selecting a hypothesis different from the one selected by standard MBR using all the samples ($N=64$).}
    \label{fig:sum-cap}
\end{figure*}

\subsection{Machine Translation}
\label{sec:nmt}

We evaluate the performance on machine translation tasks using WMT'21 test dataset. We use German-English (De-En) and Russian-English (Ru-En) language pairs.
We use the WMT 21 X-En model and M2M100 418M model to sample sequences for both language pairs \cite{tran-etal-2021-facebook,10.5555/3546258.3546365}.
We load the WMT 21 X-En model in 4-bit precision to reduce the GPU memory consumption.
We use COMET-20 as the utility function and the evaluation metric \cite{rei-etal-2020-unbabels}. 
We use the BLEU score as a coarse utility function of C2F.
As a reference, the COMET-20 scores of the greedy MAP decoding are 56.84 for De-En and 54.71 for Ru-En.

\paragraph{AMBR is on par with Oracle CBP.} Figure \ref{fig:mt} shows the results with varying evaluation budgets with a fixed number of samples ($N=64, 128$) using the WMT 21 X-En model. We observe that AMBR achieves the best COMET score and the error rate compared to the others. The error rate is the ratio of selecting a hypothesis different from the standard MBR using all 128 (64) samples. It achieves almost the same score as the standard MBR with all samples within $1/4-1/8$ number of evaluations, resulting in $4-8$ times speed up compared to standard MBR. 
We observe qualitatively the same result on the M2M100 model (see Appendix \ref{sec:m2m100}).
Additional evaluations on WMT'21 En-De and En-Ru are described in Appendix \ref{sec:noneng}.

\paragraph{AMBR scales with the number of samples given enough budget.} 
To evaluate the scalability of AMBR on the number of samples, we evaluate the COMET scores with varying numbers of samples with a fixed amount of evaluation budgets using the M2M100 418M model.
Figure \ref{fig:samples} shows the COMET scores with varying numbers of samples with a fixed amount of evaluation budgets on De-En. 
The COMET score of AMBR scales with the number of samples if and only if the number of evaluations is large enough. This is to be expected as Lemma \ref{lm:ambr} only holds when the budget is large enough.
The same trend is observed on Ru-En (Appendix \ref{sec:samples-ruen}).



\subsection{Text Summarization}
\label{sec:summarization}

We evaluate the performance of AMBR on text summarization tasks using SAMSum \cite{gliwa-etal-2019-samsum} and XSum dataset \cite{narayan-etal-2018-dont}. We use BART models fine-tuned on each dataset \cite{lewis-etal-2020-bart}. 
We use InfoLM \cite{colombo2022infolm} with the Fisher-Rao distance \cite{rao1987differential} as a utility function as it is shown to have a high correlation with human judgment on text summarization tasks.
We generate $N=64$ samples as a candidate set for each input.
Following \citet{eikema-aziz-2022-sampling}, we use the F1 score of the unigram as a coarse utility function of C2F.

The results are summarized in Figure \ref{fig:sum-cap}.
Despite AMBR reduces the error rate significantly (Figure \ref{fig:samsum-err} and \ref{fig:xsum-err}), it only slightly improves upon standard MBR with respect to InfoLM and ROUGE-L score (Figure \ref{fig:samsum}, \ref{fig:samsum-rouge}, \ref{fig:xsum}, and \ref{fig:xsum-rouge}). We speculate that this is because many of the top-scoring samples are similar in quality measured by InfoLM and ROUGE-L.


\paragraph{C2F can surpass the score of standard MBR under conditions} Interestingly, we observe that C2F surpasses the performance of standard MBR with all samples on ROUGE-L for XSum dataset. We speculate that C2F may improve upon MBR because it effectively ensembles two utility functions. Because the F1 score of the unigram may be more aligned to ROUGE-L score than InfoLM is, it can pick sentences favored by ROUGE-L metric. As such, C2F can not only speed up the computation of the MBR objective but also improve the alignment to the target metric.



\subsection{Image Captioning}
\label{sec:captioning}

We evaluate the performance of AMBR on image captioning task using MS COCO dataset \cite{lin2014microsoft}. We use BLIP-2 \cite{pmlr-v202-li23q} with Flan T5-xl \cite{chung2022scaling} fine-tuned for MS COCO loaded in 4-bit  precision. 
We use a cosine similarity of the textual CLIP embeddings as the utility function \cite{radford2021learning,hessel-etal-2021-clipscore}. 
We use RefCLIPScore and BLEU as an evaluation metric \cite{hessel-etal-2021-clipscore,papineni-etal-2002-bleu}.
We generate $N=64$ samples for each image.
We use the F1 score of the unigram as a coarse utility function of C2F.

The empirical result is shown in Figure \ref{fig:sum-cap}. AMBR achieves roughly $4$ to $8$ times speed-up compared to MBR with a marginal drop in RefCLIPScore and BLEU score (Figure \ref{fig:mscoco} and \ref{fig:mscoco-bleu}).
We observe C2F to improve upon standard MBR with respect to BLEU score (Figure \ref{fig:mscoco-bleu}). As in text summarization (Section \ref{sec:summarization}), We speculate that this is because the F1 score has a better alignment with the BLEU score than the CLIP embeddings so that the coarse utility function is effectively serving as another utility function.

\section{Related Work}

MBR has been investigated in many NLP tasks including parsing \cite{goodman-1996-parsing}, speech recognition \cite{goel2000minimum}, bilingual word alignment \cite{kumar-byrne-2002-minimum}, and machine translation \cite{kumar-byrne-2004-minimum}.
MBR has recently gained attention in machine translation as a decision rule as a method to overcome some of the biases of MAP decoding in NMT \cite{eikema-aziz-2020-map,muller-sennrich-2021-understanding,eikema-aziz-2022-sampling}. 

\citet{freitag-etal-2022-high} and \citet{fernandes-etal-2022-quality} show that using neural-based utility functions such as BLEURT \cite{sellam-etal-2020-bleurt,pu-etal-2021-learning} and COMET \cite{rei-etal-2020-unbabels,rei-etal-2022-comet} rather than lexical overlap metrics (e.g. BLEU) further improves MBR.


CSH \cite{baharav2019ultra} is not the only algorithm proposed to solve the medoid identification problem. 
There are several other algorithms to solve the medoid identification \cite{wang2006fast,okamoto2008ranking,bagaria2018medoids}.
We pick to use CSH as it has the best theoretical performance.

Algorithms to solve the problem of identifying the best option out of the candidates with a budget constraint (fixed-budget best-arm identification problems) are known to be highly sensitive to the choice of the hyperparameters if they have ones \cite{pmlr-v49-carpentier16,JMLR:v17:kaufman16a}. In fact, we observe that the effectiveness of CBP hinges on the appropriate selection of hyperparameters, given each budget constraint.

\section{Conclusions}

We propose Adaptive Minimum Bayes-Risk (AMBR) decoding, a hyperparameter-free algorithm for efficient MBR decoding. AMBR considers the problem of computing the MBR objective as the medoid identification problem and uses the known best algorithm to solve it. 
The strength of the AMBR is that it doesn't need a development set to tune the set of hyperparameters. AMBR automatically computes the strategy on the fly given the budget specified by the user.


Experimental result shows that the performance of AMBR is on par with CBP with hyperparameters picked by an Oracle on machine translation tasks.
AMBR outperforms CBP on text summarization and image captioning tasks, using the same set of hyperparameters as in machine translation tasks for CBP. We speculate that CBP requires a different set of hyperparameters for each task to perform on par with AMBR.



We believe that AMBR will be a practical choice for future MBR decoding because of its applicability and significant performance improvements.



\section{Limitations}


Even with the improvement, AMBR is still many times more costly to run than beam search. 

Using Eq.~\eqref{eq:guaranteed}, the computational complexity of the evaluation of the utility function of AMBR is $O(N \log N \cdot U)$ to achieve the theoretical guarantee. 
This is still larger than the complexity of generation which is $O(N \cdot G)$. Therefore, the evaluation procedure is still the bottleneck of MBR to scale with the number of samples. 

Although our focus is on reducing the computation of the utility function of MBR decoding, it is not the only way to speed up the text generation.
\citet{finkelstein2023mbr} shows that by self-training a machine translation model by its own MBR-decoded output, it can improve the performance of more efficient decoding methods such as beam search.
\citet{yang2023direct} proposes the use of Direct Preference Optimization \cite{rafailov2023direct} to train the model to learn the ranking of the sequences according to the MBR objective.
\citet{foks2023towards} shows that by training a model to predict the Monte Carlo estimate of the Bayes risk, we can directly estimate the Bayes risk using the trained model without running Monte Carlo estimation, resulting in $O(N \cdot G + N \cdot U')$ where $U'$ is the inference time of the trained model. 

We measure the number of evaluations of the utility function as a metric of efficiency. Practically, the computation of the utility function is not linear to the number of calls. One can optimize the implementation by batching and caching the computation effectively. For example, the sentence embeddings of embedding-based utility functions such as COMET can be cached to significantly speed up the computation of the utility \cite{amrhein-sennrich-2022-identifying,cheng-vlachos-2023-faster}.

The paper focuses on how to effectively use the given budget and lacks a discussion on what to set the budget to. \citet{baharav2019ultra} suggests the doubling trick \cite{besson2018doubling} to find the appropriate budget size. That is, we run the algorithm with a certain budget $T$, and then double the budget to $2T$ and rerun the algorithm. If the two answers are the same, then we output it. Because the probability of selecting the same incorrect answer twice in a row is very low, it is likely to be the best hypothesis. Empirical evaluation of the strategies to decide the budget size is future work.

The other question is on what to set the number of samples to. Figure \ref{fig:samples} shows that having too many samples is not necessarily beneficial when the evaluation budget is too small. Finding the optimal number of samples given a budget on computation is an open question.



We consider the Monte Carlo estimate $\vh^{\mathrm{MC}}$ as the target objective function to compute. Evaluation of AMBR using other objective functions such as model-based estimate \cite{jinnai2023modelbased} is future work.

Although AMBR is based on the best algorithm known to solve the medoid identification problem, it does not use any task-dependent knowledge to speed up the algorithm. One may exploit the domain knowledge of the task to further improve upon it (e.g. reference aggregation; \citealp{vamvas2024lineartime}).



\section*{Acknowledgements}

We thank all the reviewers for their constructive comments throughout the manuscript review.
Kaito Ariu's research is supported by JSPS KAKENHI Grant No. 23K19986.


\appendix


\section{Minimum Bayes Risk Decoding with Reward Model for Alignment}
\label{sec:reward}

Reference aggregation is not applicable using an utility function that is not aggregatable. In the following experiment, we show an instance of an utility function that is practically useful but not aggregatable.

We evaluate the performance of MBR and its variants on Alpaca Eval dataset \cite{alpaca_eval}.
The task is to generate a response to a human query that follows the human preference.
One of the popular decoding strategy for LLMs is best-of-n strategy \cite{NEURIPS2020_1f89885d,nakano2022webgpt}. Best-of-n generates multiple outputs and simply picks the output with the highest reward value according to a reward function $R$ that is trained to predict the human preference:
\begin{equation}
    \vh^{\mathrm{bon}} = \argmax_{\vh \in \CandH} R(\vy).
\end{equation}
MBR decoding is also shown to be an efficient strategy on text generation tasks using large language models (LLMs) \cite{li2024agents}.\footnote{MBR decoding is called Sampling-and-voting (Algorithm 1) in \cite{li2024agents}.}
We compare the performance of epsilon sampling, best-of-n, MBR without using a reward function \cite{li2024agents}, MBR with reference aggregation without using a reward function (RA-MBR) \cite{vamvas2024lineartime}, MBR using a reward function, and AMBR using a reward function.
We implement MBR using a reward function as follows:
\begin{equation}
    \vh^{\mathrm{reward}} = \argmax_{\vh \in \CandH} \frac{1}{|\RefH|} \sum_{\vy \in \RefH} u(\vh, \vy) \cdot R(\vy).
\end{equation}
Note that $\vh^{\mathrm{reward}}$ is not immediately aggregatable as most of the state-of-the-art reward functions are based on transformer architecture where the input is a sequence of token embeddings instead of a sentence embedding. Thus, RA-MBR is not directly applicable when combined with a reward function.
We use Mistral-7B-Instruct-v0.1 \cite{jiang2023mistral} as the text generation model.
We generate 128 samples with epsilon sampling with $\epsilon=0.01$ for best-of-n and MBRs. 
We use sentence BERT \cite{reimers-gurevych-2019-sentence} as the utility function $u$.
We compute the embedding of each output using the sentence BERT and compute the cosine similarity of each pair of outputs. 
We use \textsc{all-mpnet-base-v2} model as it has shown to be one of the most effective in sentence embedding tasks.\footnote{\url{https://huggingface.co/sentence-transformers/all-mpnet-base-v2}}
We use SteamSHP-Large as a reward function \cite{pmlr-v162-ethayarajh22a}.
The budget of AMBR is set to 10000.
The output is evaluated using an OASST reward model as the gold reference \cite{kopf2023openassistant}.\footnote{\url{OpenAssistant/reward-model-deberta-v3-large-v2}}
We use OASST as it is shown to be one of the most accurate reward model in prior work \cite{touvron2023llama,cui2023ultrafeedback}.

\begin{figure}
    \centering
    \includegraphics[width=\columnwidth]{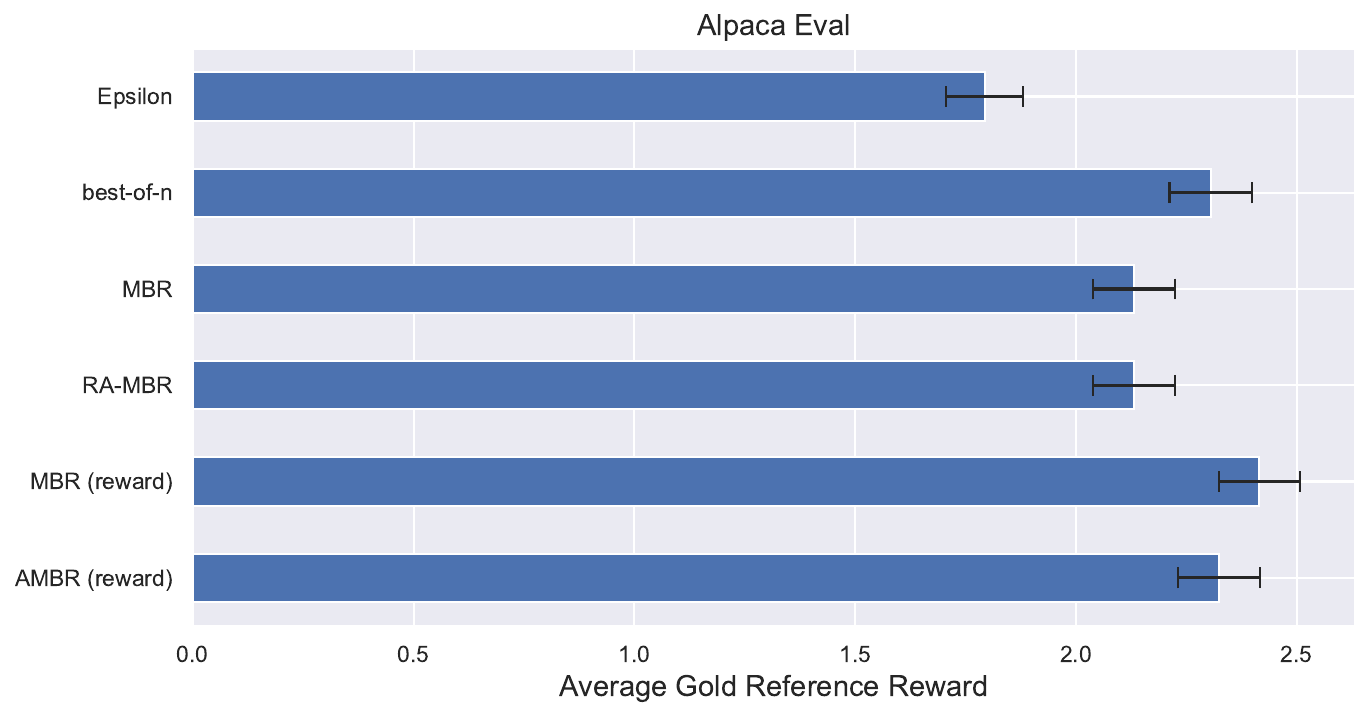}
    \caption{Average reward according to OASST (gold reference reward) on Alpaca Eval dataset.}
    \label{fig:reward}
\end{figure}

Figure \ref{fig:reward} is the summary of the reward scores. While MBR and RA-MBR without using a reward model has lower score than best-of-n, MBR with a reward function has higher score than best-of-n.
RA-MBR achieves mostly the same score as MBR as the utility function is a cosine similarity of the sentence embedding itself. Thus, linear aggregation of the references results in exactly the mean of the references in the embedding space. Still, because it does not use the reward function, its score is lower than best-of-n and MBRs with reward functions.

The analysis shows that in this setting, non-aggregatable utility function has a potential to achieve higher performance than aggregatable one, and thus reference aggregation is not applicable but AMBR is.

\section{Formulation of Medoid Identification Problem}
\label{sec:medoid}

We show that the Eq.~\eqref{eq:medoid} represents the same class of problem as the standard formulation of the medoid identification problem where $X=Y$ is assumed.
Let $(d, X, Y)$ be an instance of generalized medoid identification problem (Eq.~\ref{eq:medoid}): 
\begin{equation}
    \vx^* = \argmin_{\vx \in X} \sum_{\vy \in Y} d(\vx, \vy). \nonumber
\end{equation}
Let $X' = X \cup Y$ and $d'$ as follows:
\begin{equation}
    d'(\vx, \vy) = \begin{cases}
    \infty       & \text{if $\vx \notin X$} \\
    0            & \text{if $\vx \in X \land \vy \notin Y$} \\
    d(\vx, \vy)  & \text{otherwise.} \\
  \end{cases}
\end{equation}
Then, $(d', X', X')$ returns the same solution as $(d, X, Y)$.
Thus, Eq.~\eqref{eq:medoid} represents the same class of problem as the standard formulation of the medoid identification problem where $X=Y$ is assumed.

\section{Walltime}
\label{apx:walltime}
We describe the wall clock time of AMBR in Table~\ref{tab:walltime}. We run with COMET as the utility function on WMT'21 De-En. 
We set $r=1$ and $\alpha=0.99$ for CBP. We set the budget size of AMBR to $1/8$ of the number of evaluations of the standard MBR.
We cache the sentence embedding on computing COMET to speed up the utility function calls \cite{amrhein-sennrich-2022-identifying,cheng-vlachos-2023-faster}. 
The batch size for generating $\CandH$ is set to $4$. We set the batch size for the computation of COMET to $64$.
In our code base, we compute the generation probability of the sampled sequence when generating $\CandH$, which is also included in the wall time in the Table.
Note that the wall clock time depends on various factors including the code base and the hardware. 
All experiments are conducted using an NVIDIA A2 GPU.

\begin{table}
    \centering
\adjustbox{max width=0.98\columnwidth}{
    \begin{tabular}{cccc}
    \toprule
        & \multicolumn{3}{c}{Run time (seconds)} \\
        \cmidrule(l){2-4}
                      & Standard & CBP & AMBR \\\midrule
    Generate $\CandH$ & 41964 & 41964 & 41964 \\
    Compute utility  & 123471 & 52219 & 50740 \\
    \bottomrule
    \end{tabular}
}
    \caption{Summary of wall clock time of AMBR on 1000 inputs with $N=128$ for WMT'21 De-En. All experiments are run on an NVIDIA A2 GPU.}
    \label{tab:walltime}
\end{table}

\section{Hyperparameters for Confidence-Based Pruning}
\label{sec:cbp-params}

The performance of CBP with varying hyperparameters is present in Table \ref{tab:cbp-params} (machine translation), Table \ref{tab:cbp-params-sum} (text summarization), and Table \ref{tab:cbp-params-mscoco} (image captioning). The average score over five runs is reported.
Smaller $r_0$ and $\alpha$ tend to achieve higher COMET scores when the budget is small and larger $r_0$ and $\alpha$ achieve higher scores when the budget is large enough.

\begin{table*}
    \centering
\adjustbox{max width=0.95\textwidth}{
\begin{tabular}{ll|rr|rr|rr|rr|rr|rr}
\toprule
\multicolumn{2}{c|}{\#Evaluations} & \multicolumn{2}{c|}{Mean} & \multicolumn{2}{c|}{1/32} & \multicolumn{2}{c|}{1/16} & \multicolumn{2}{c|}{1/8} & \multicolumn{2}{c|}{1/4} & \multicolumn{2}{c}{1/2} \\
$r_0$ & $\alpha$ & COMET & Rank & COMET & Rank & COMET & Rank & COMET & Rank & COMET & Rank & COMET & Rank \\
\midrule\midrule
\multicolumn{14}{c}{WMT'21 De-En ($N=128$)} \\
\midrule
1 & 0.80 & 64.00 & 2 & 62.14 & 2 & 64.21 & 2 & 64.56 & 1 & 64.56 & 2 & 64.55 & 9 \\
1 & 0.90 & 63.76 & 4 & 61.43 & 4 & 63.80 & 5 & 64.50 & 5 & 64.53 & 11 & 64.55 & 8 \\
1 & 0.99 & 63.33 & 8 & 60.36 & 7 & 63.02 & 10 & 64.22 & 10 & 64.47 & 13 & 64.58 & 1 \\
2 & 0.80 & 63.85 & 3 & 61.60 & 3 & 64.08 & 3 & 64.50 & 3 & 64.54 & 6 & 64.55 & 13 \\
2 & 0.90 & 63.72 & 5 & 61.23 & 5 & 63.78 & 6 & 64.43 & 7 & 64.59 & 1 & 64.57 & 3 \\
2 & 0.99 & 63.37 & 6 & 60.50 & 6 & 63.18 & 8 & 64.11 & 12 & 64.52 & 12 & 64.56 & 5 \\
4 & 0.80 & 63.35 & 7 & 59.33 & 10 & 63.83 & 4 & 64.48 & 6 & 64.55 & 5 & 64.55 & 11 \\
4 & 0.90 & 63.31 & 9 & 59.19 & 13 & 63.75 & 7 & 64.50 & 4 & 64.54 & 7 & 64.56 & 4 \\
4 & 0.99 & 63.17 & 10 & 59.47 & 8 & 63.07 & 9 & 64.20 & 11 & 64.54 & 8 & 64.55 & 10 \\
8 & 0.80 & 63.12 & 11 & 59.33 & 9 & 62.67 & 13 & 64.50 & 2 & 64.56 & 3 & 64.55 & 12 \\
8 & 0.90 & 63.10 & 12 & 59.27 & 11 & 62.80 & 12 & 64.32 & 9 & 64.55 & 4 & 64.56 & 6 \\
8 & 0.99 & 63.06 & 13 & 59.22 & 12 & 62.89 & 11 & 64.08 & 13 & 64.54 & 9 & 64.58 & 2 \\
\midrule
\multicolumn{2}{l|}{AMBR} & 64.31 & 1 & 63.81 & 1 & 64.23 & 1 & 64.42 & 8 & 64.53 & 10 & 64.55 & 7 \\
\midrule\midrule
\multicolumn{14}{c}{WMT'21 Ru-En ($N=128$)} \\
\midrule
1 & 0.80 & 63.33 & 2 & 61.73 & 2 & 63.39 & 3 & 63.82 & 3 & 63.86 & 5 & 63.87 & 10 \\
1 & 0.90 & 63.20 & 4 & 61.20 & 4 & 63.20 & 5 & 63.84 & 2 & 63.88 & 3 & 63.88 & 4 \\
1 & 0.99 & 62.83 & 6 & 60.17 & 7 & 62.79 & 8 & 63.45 & 13 & 63.85 & 6 & 63.90 & 1 \\
2 & 0.80 & 63.23 & 3 & 61.20 & 3 & 63.42 & 2 & 63.84 & 1 & 63.83 & 8 & 63.87 & 11 \\
2 & 0.90 & 63.14 & 5 & 60.93 & 5 & 63.27 & 4 & 63.80 & 5 & 63.83 & 9 & 63.87 & 8 \\
2 & 0.99 & 62.80 & 7 & 60.27 & 6 & 62.66 & 10 & 63.45 & 12 & 63.72 & 13 & 63.87 & 9 \\
4 & 0.80 & 62.74 & 9 & 58.98 & 12 & 63.17 & 6 & 63.77 & 6 & 63.92 & 1 & 63.87 & 7 \\
4 & 0.90 & 62.75 & 8 & 59.25 & 10 & 63.02 & 7 & 63.70 & 9 & 63.88 & 2 & 63.89 & 3 \\
4 & 0.99 & 62.65 & 10 & 59.27 & 9 & 62.68 & 9 & 63.58 & 10 & 63.81 & 10 & 63.90 & 2 \\
8 & 0.80 & 62.54 & 13 & 58.95 & 13 & 62.31 & 12 & 63.75 & 8 & 63.81 & 11 & 63.87 & 12 \\
8 & 0.90 & 62.57 & 12 & 59.04 & 11 & 62.30 & 13 & 63.76 & 7 & 63.85 & 7 & 63.88 & 6 \\
8 & 0.99 & 62.63 & 11 & 59.39 & 8 & 62.46 & 11 & 63.57 & 11 & 63.86 & 4 & 63.88 & 5 \\
\midrule
\multicolumn{2}{l|}{AMBR} & 63.61 & 1 & 63.12 & 1 & 63.51 & 1 & 63.80 & 4 & 63.80 & 12 & 63.80 & 13 \\
\midrule\midrule
\multicolumn{14}{c}{WMT'21 De-En ($N=64$)} \\
\midrule
1 & 0.80 & 61.06 & 2 & 52.54 & 2 & 61.35 & 2 & 63.34 & 3 & 64.01 & 2 & 64.04 & 11 \\
1 & 0.90 & 60.68 & 3 & 51.76 & 3 & 60.52 & 5 & 63.17 & 5 & 63.90 & 8 & 64.07 & 2 \\
1 & 0.99 & 59.57 & 9 & 47.93 & 12 & 59.48 & 6 & 62.67 & 9 & 63.70 & 12 & 64.07 & 3 \\
2 & 0.80 & 60.18 & 4 & 48.26 & 7 & 61.05 & 3 & 63.55 & 2 & 63.97 & 3 & 64.07 & 4 \\
2 & 0.90 & 59.97 & 5 & 48.00 & 11 & 60.53 & 4 & 63.28 & 4 & 63.94 & 6 & 64.08 & 1 \\
2 & 0.99 & 59.59 & 8 & 48.26 & 8 & 59.13 & 7 & 62.79 & 8 & 63.72 & 11 & 64.04 & 10 \\
4 & 0.80 & 59.67 & 7 & 48.43 & 5 & 58.79 & 9 & 63.16 & 6 & 63.94 & 5 & 64.05 & 7 \\
4 & 0.90 & 59.69 & 6 & 48.65 & 4 & 58.64 & 13 & 63.13 & 7 & 63.96 & 4 & 64.06 & 5 \\
4 & 0.99 & 59.40 & 11 & 48.11 & 9 & 58.65 & 11 & 62.43 & 10 & 63.77 & 10 & 64.03 & 12 \\
8 & 0.80 & 59.43 & 10 & 48.28 & 6 & 58.69 & 10 & 62.26 & 11 & 63.90 & 9 & 64.04 & 9 \\
8 & 0.90 & 59.35 & 12 & 48.02 & 10 & 58.64 & 12 & 62.10 & 13 & 63.90 & 7 & 64.06 & 6 \\
8 & 0.99 & 59.25 & 13 & 47.24 & 13 & 59.06 & 8 & 62.24 & 12 & 63.65 & 13 & 64.05 & 8 \\
\midrule
\multicolumn{2}{l|}{AMBR} & 63.29 & 1 & 61.49 & 1 & 63.10 & 1 & 63.80 & 1 & 64.02 & 1 & 64.03 & 13 \\
\midrule\midrule
\multicolumn{14}{c}{WMT'21 Ru-En ($N=64$)} \\
\midrule
1 & 0.80 & 60.63 & 2 & 53.30 & 2 & 60.65 & 3 & 62.79 & 3 & 63.18 & 4 & 63.23 & 5 \\
1 & 0.90 & 60.44 & 3 & 52.63 & 3 & 60.50 & 4 & 62.70 & 5 & 63.13 & 7 & 63.24 & 2 \\
1 & 0.99 & 59.13 & 10 & 48.31 & 11 & 58.99 & 7 & 62.12 & 9 & 62.99 & 13 & 63.21 & 11 \\
2 & 0.80 & 59.68 & 4 & 48.00 & 13 & 60.93 & 2 & 63.02 & 1 & 63.19 & 3 & 63.24 & 3 \\
2 & 0.90 & 59.58 & 5 & 48.58 & 7 & 60.18 & 5 & 62.68 & 6 & 63.23 & 1 & 63.22 & 9 \\
2 & 0.99 & 59.20 & 8 & 48.40 & 9 & 59.31 & 6 & 62.07 & 10 & 63.01 & 12 & 63.22 & 10 \\
4 & 0.80 & 59.26 & 7 & 48.32 & 10 & 58.82 & 9 & 62.72 & 4 & 63.20 & 2 & 63.23 & 4 \\
4 & 0.90 & 59.27 & 6 & 48.70 & 4 & 58.87 & 8 & 62.45 & 7 & 63.09 & 10 & 63.23 & 6 \\
4 & 0.99 & 59.17 & 9 & 48.65 & 5 & 58.71 & 11 & 62.15 & 8 & 63.10 & 9 & 63.22 & 8 \\
8 & 0.80 & 59.03 & 13 & 48.19 & 12 & 58.65 & 13 & 61.92 & 11 & 63.14 & 6 & 63.25 & 1 \\
8 & 0.90 & 59.11 & 11 & 48.61 & 6 & 58.78 & 10 & 61.82 & 13 & 63.11 & 8 & 63.21 & 12 \\
8 & 0.99 & 59.06 & 12 & 48.52 & 8 & 58.70 & 12 & 61.87 & 12 & 63.02 & 11 & 63.20 & 13 \\
\midrule
\multicolumn{2}{l|}{AMBR} & 62.53 & 1 & 60.85 & 1 & 62.46 & 1 & 62.96 & 2 & 63.17 & 5 & 63.22 & 7 \\
\bottomrule
\end{tabular}
}
    \caption{Evaluation of confidence-based pruning (CBP) with varying hyperparameters. $r_0$ is the number of references at the first iteration. $\alpha$ is the threshold of the win rate on pruning. The average COMET-20 score over five runs is reported. Rank denotes the rank of the average COMET-20 score over a set of runs of CBP and AMBR. Mean column reports the average COMET score over 1/32, 1/16, 1/8, 1/4, 1/2.}
    \label{tab:cbp-params}
\end{table*}

\begin{table*}
    \centering
\adjustbox{max width=\textwidth}{
\begin{tabular}{ll|rr|rr|rr|rr|rr|rr}
\toprule
\multicolumn{2}{c|}{\#Evaluations} & \multicolumn{2}{c|}{Mean} & \multicolumn{2}{c|}{1/32} & \multicolumn{2}{c|}{1/16} & \multicolumn{2}{c|}{1/8} & \multicolumn{2}{c|}{1/4} & \multicolumn{2}{c}{1/2} \\
$r_0$ & $\alpha$ & InfoLM & Rank & InfoLM & Rank & InfoLM & Rank & InfoLM & Rank & InfoLM & Rank & InfoLM & Rank \\
\midrule\midrule
\multicolumn{14}{c}{SAMSum ($N=64$)} \\
\midrule
1 & 0.80 & 1.864 & 4 & 1.982 & 13 & 1.896 & 2 & 1.850 & 4 & 1.802 & 2 & 1.792 & 3 \\
1 & 0.90 & 1.868 & 8 & 1.976 & 12 & 1.902 & 4 & 1.856 & 7 & 1.812 & 7 & 1.794 & 8 \\
1 & 0.99 & 1.870 & 10 & 1.960 & 8 & 1.909 & 5 & 1.863 & 11 & 1.821 & 11 & 1.798 & 11 \\
2 & 0.80 & 1.860 & 2 & 1.955 & 2 & 1.902 & 3 & 1.845 & 2 & 1.803 & 3 & 1.793 & 6 \\
2 & 0.90 & 1.866 & 6 & 1.958 & 5 & 1.909 & 6 & 1.856 & 6 & 1.813 & 8 & 1.793 & 5 \\
2 & 0.99 & 1.871 & 12 & 1.960 & 9 & 1.916 & 11 & 1.858 & 8 & 1.821 & 12 & 1.797 & 10 \\
4 & 0.80 & 1.864 & 3 & 1.961 & 10 & 1.915 & 10 & 1.847 & 3 & 1.806 & 4 & 1.793 & 4 \\
4 & 0.90 & 1.865 & 5 & 1.956 & 3 & 1.912 & 7 & 1.854 & 5 & 1.811 & 6 & 1.793 & 7 \\
4 & 0.99 & 1.872 & 13 & 1.957 & 4 & 1.922 & 13 & 1.860 & 10 & 1.822 & 13 & 1.801 & 13 \\
8 & 0.80 & 1.867 & 7 & 1.961 & 11 & 1.916 & 12 & 1.859 & 9 & 1.809 & 5 & 1.792 & 1 \\
8 & 0.90 & 1.869 & 9 & 1.958 & 6 & 1.913 & 9 & 1.864 & 13 & 1.816 & 9 & 1.795 & 9 \\
8 & 0.99 & 1.870 & 11 & 1.959 & 7 & 1.912 & 8 & 1.864 & 12 & 1.818 & 10 & 1.799 & 12 \\
\midrule
\multicolumn{2}{l|}{AMBR} & 1.823 & 1 & 1.902 & 1 & 1.820 & 1 & 1.802 & 1 & 1.796 & 1 & 1.792 & 2 \\
\midrule\midrule
\multicolumn{14}{c}{XSum ($N=64$)} \\
\midrule
1 & 0.80 & 1.954 & 3 & 2.069 & 13 & 1.997 & 3 & 1.935 & 2 & 1.889 & 2 & 1.880 & 4 \\
1 & 0.90 & 1.961 & 8 & 2.066 & 12 & 1.998 & 4 & 1.952 & 7 & 1.904 & 7 & 1.882 & 8 \\
1 & 0.99 & 1.964 & 12 & 2.057 & 10 & 2.006 & 9 & 1.961 & 13 & 1.910 & 9 & 1.888 & 13 \\
2 & 0.80 & 1.952 & 2 & 2.056 & 9 & 1.993 & 2 & 1.943 & 3 & 1.891 & 3 & 1.878 & 1 \\
2 & 0.90 & 1.959 & 6 & 2.052 & 3 & 2.003 & 5 & 1.950 & 5 & 1.909 & 8 & 1.881 & 6 \\
2 & 0.99 & 1.963 & 11 & 2.053 & 8 & 2.008 & 12 & 1.955 & 10 & 1.912 & 10 & 1.886 & 11 \\
4 & 0.80 & 1.956 & 4 & 2.057 & 11 & 2.005 & 8 & 1.943 & 4 & 1.893 & 4 & 1.880 & 3 \\
4 & 0.90 & 1.960 & 7 & 2.053 & 7 & 2.008 & 11 & 1.954 & 8 & 1.903 & 6 & 1.882 & 7 \\
4 & 0.99 & 1.963 & 10 & 2.052 & 5 & 2.006 & 10 & 1.951 & 6 & 1.920 & 13 & 1.885 & 10 \\
8 & 0.80 & 1.957 & 5 & 2.052 & 2 & 2.004 & 6 & 1.954 & 9 & 1.896 & 5 & 1.879 & 2 \\
8 & 0.90 & 1.962 & 9 & 2.052 & 4 & 2.004 & 7 & 1.956 & 11 & 1.914 & 11 & 1.883 & 9 \\
8 & 0.99 & 1.965 & 13 & 2.053 & 6 & 2.012 & 13 & 1.960 & 12 & 1.915 & 12 & 1.886 & 12 \\
\midrule
\multicolumn{2}{l|}{AMBR} & 1.913 & 1 & 1.990 & 1 & 1.913 & 1 & 1.892 & 1 & 1.886 & 1 & 1.881 & 5 \\
\bottomrule
\end{tabular}
}
    \caption{Evaluation of confidence-based pruning (CBP) with varying hyperparameters on SAMSum and XSum. $r_0$ is the number of references at the first iteration. $\alpha$ is the threshold of the win rate on pruning. The average InfoLM over five runs is reported. Rank denotes the rank of the average score over a set of runs of CBP and AMBR. Mean column reports the average InfoLM score over 1/32, 1/16, 1/8, 1/4, 1/2.}
    \label{tab:cbp-params-sum}
\end{table*}

\begin{table*}
    \centering
\adjustbox{max width=\textwidth}{
\begin{tabular}{ll|rr|rr|rr|rr|rr|rr}
\toprule
\multicolumn{2}{c|}{\#Evaluations} & \multicolumn{2}{c|}{Mean} & \multicolumn{2}{c|}{1/32} & \multicolumn{2}{c|}{1/16} & \multicolumn{2}{c|}{1/8} & \multicolumn{2}{c|}{1/4} & \multicolumn{2}{c}{1/2} \\
$r_0$ & $\alpha$ & RCLIP & Rank & RCLIP & Rank & RCLIP & Rank & RCLIP & Rank & RCLIP & Rank & RCLIP & Rank \\
\midrule\midrule
\multicolumn{14}{c}{MS COCO ($N=64$)} \\
\midrule
1 & 0.80 & 39.27 & 3 & 38.09 & 13 & 39.10 & 2 & 39.60 & 2 & 39.76 & 2 & 39.81 & 4 \\
1 & 0.90 & 39.22 & 7 & 38.10 & 12 & 38.99 & 5 & 39.49 & 7 & 39.72 & 8 & 39.81 & 2 \\
1 & 0.99 & 39.21 & 10 & 38.28 & 5 & 38.86 & 10 & 39.44 & 11 & 39.69 & 11 & 39.78 & 12 \\
2 & 0.80 & 39.29 & 2 & 38.23 & 8 & 39.08 & 3 & 39.58 & 3 & 39.75 & 4 & 39.81 & 3 \\
2 & 0.90 & 39.26 & 4 & 38.23 & 10 & 39.01 & 4 & 39.56 & 5 & 39.73 & 7 & 39.80 & 6 \\
2 & 0.99 & 39.22 & 8 & 38.33 & 2 & 38.89 & 6 & 39.42 & 13 & 39.67 & 12 & 39.78 & 11 \\
4 & 0.80 & 39.26 & 5 & 38.29 & 3 & 38.88 & 7 & 39.56 & 4 & 39.75 & 3 & 39.81 & 1 \\
4 & 0.90 & 39.24 & 6 & 38.28 & 4 & 38.88 & 8 & 39.52 & 6 & 39.74 & 6 & 39.80 & 7 \\
4 & 0.99 & 39.19 & 12 & 38.23 & 9 & 38.81 & 13 & 39.46 & 8 & 39.69 & 10 & 39.76 & 13 \\
8 & 0.80 & 39.21 & 9 & 38.21 & 11 & 38.86 & 9 & 39.45 & 9 & 39.75 & 5 & 39.80 & 9 \\
8 & 0.90 & 39.20 & 11 & 38.24 & 7 & 38.84 & 11 & 39.44 & 10 & 39.69 & 9 & 39.81 & 5 \\
8 & 0.99 & 39.19 & 13 & 38.25 & 6 & 38.82 & 12 & 39.42 & 12 & 39.65 & 13 & 39.79 & 10 \\
\midrule
\multicolumn{2}{l|}{AMBR} & 39.65 & 1 & 39.33 & 1 & 39.61 & 1 & 39.74 & 1 & 39.78 & 1 & 39.80 & 8 \\
\bottomrule
\end{tabular}
}
    \caption{Evaluation of confidence-based pruning (CBP) with varying hyperparameters on MS COCO. $r_0$ is the number of references at the first iteration. $\alpha$ is the threshold of the win rate on pruning. The average RefCLIPScore (RCLIP) over five runs is reported. Rank denotes the rank of the average score over a set of runs of CBP and AMBR. Mean column reports the average RCLIP score over 1/32, 1/16, 1/8, 1/4, 1/2.}
    \label{tab:cbp-params-mscoco}
\end{table*}

\section{Additional Evaluations}
We describe additional experiments to evaluate the performance of the MBR decoding algorithms.

\subsection{Error Rate on WMT'21 De-En and Ru-En}

\begin{figure*}[htb]
     \centering
     \begin{subfigure}[b]{0.45\textwidth}
         \centering
         \includegraphics[width=\textwidth]{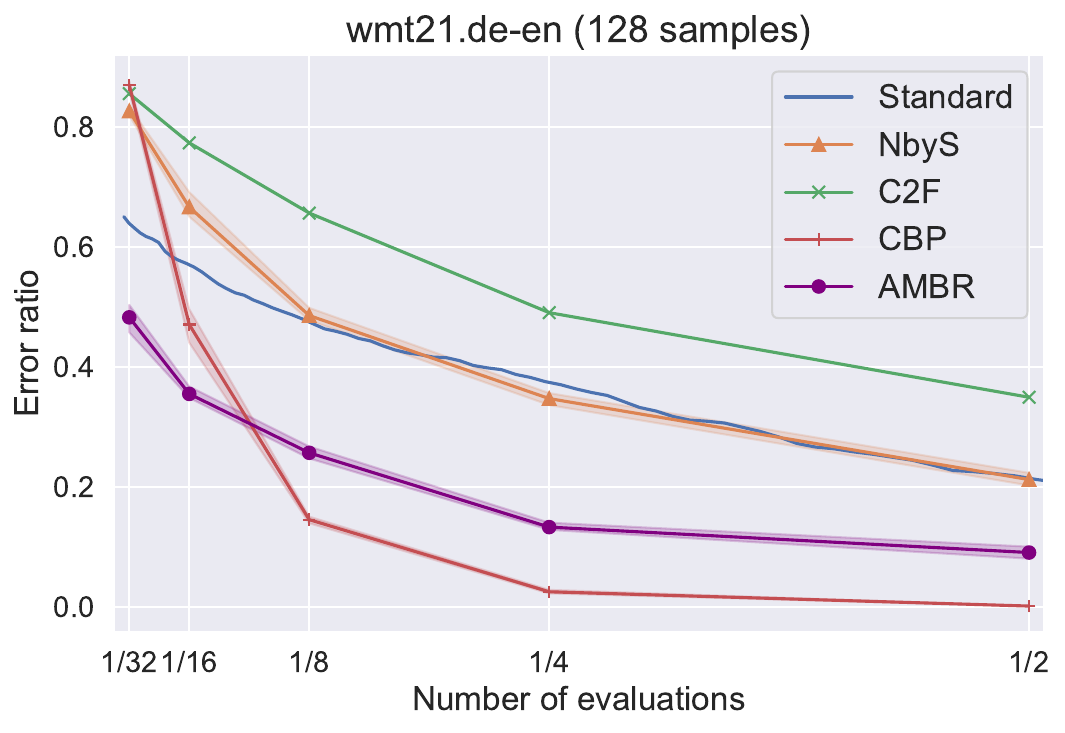}
         \caption{Error Rate $\downarrow$ (De-En, $N=128$)}
         \label{fig:deen-err}
     \end{subfigure}
     \hfill
     \begin{subfigure}[b]{0.45\textwidth}
         \centering
         \includegraphics[width=\textwidth]{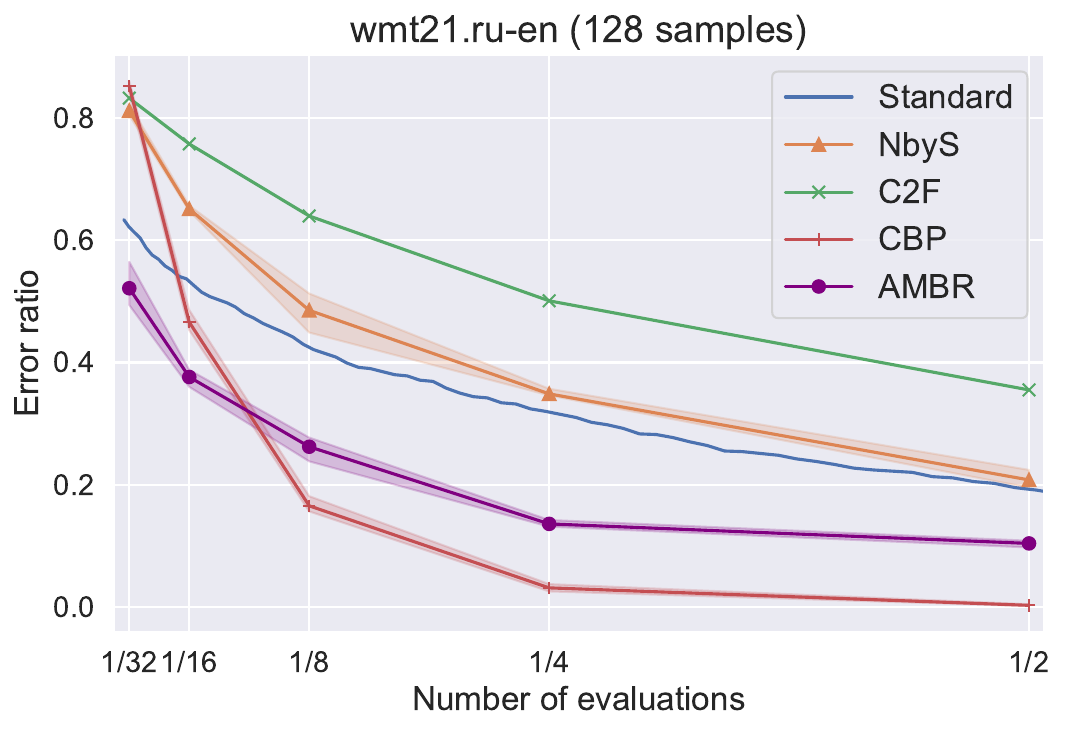}
         \caption{Error Rate $\downarrow$ (Ru-En, $N=128$)}
         \label{fig:ruen-err}
     \end{subfigure} \\
     \begin{subfigure}[b]{0.45\textwidth}
         \centering
         \includegraphics[width=\textwidth]{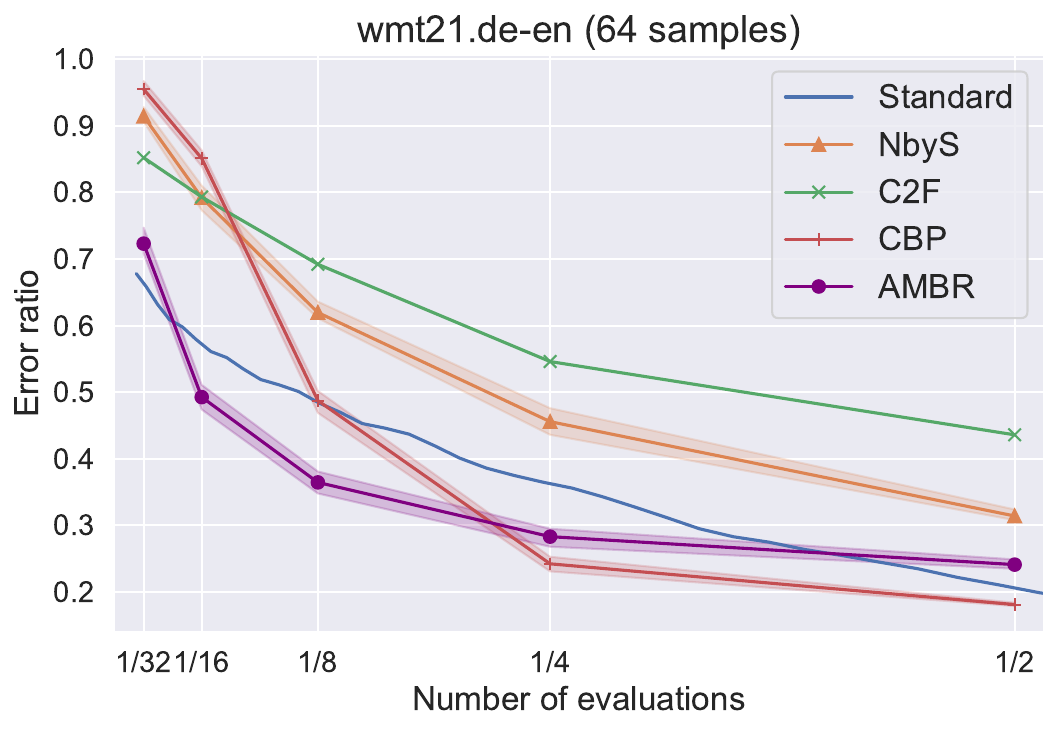}
         \caption{Error Rate $\downarrow$ (De-En, $N=64$)}
         \label{fig:deen64-err}
     \end{subfigure}
     \hfill
     \begin{subfigure}[b]{0.45\textwidth}
         \centering
         \includegraphics[width=\textwidth]{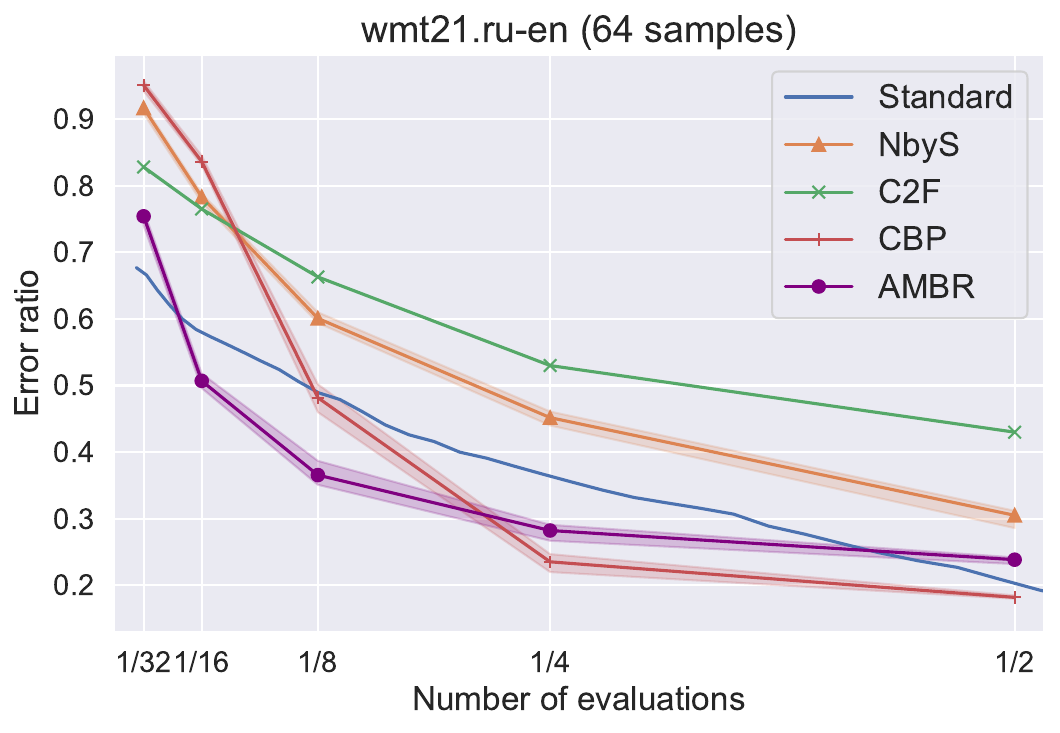}
         \caption{Error Rate $\downarrow$ (Ru-En, $N=64$)}
         \label{fig:ruen64-err}
     \end{subfigure}
    \caption{The error rate on WMT'21 De-En and Ru-En using the WMT 21 X-En model. The shaded regions show the minimum and the maximum values over five runs. The error rate is the ratio of selecting a hypothesis different from the standard MBR using all samples. The horizontal axis shows the reduction in the number of evaluations compared to the standard MBR with all samples.}
    \label{fig:mt-err}
\end{figure*}

Figure \ref{fig:mt-err} shows the error ratio for WMT'21 De-En and Ru-En.
Interestingly, although AMBR achieves a higher or equivalent COMET score to CBP (Oracle), the error ratio is higher than CBP (Oracle). This suggests that when AMBR fails to find the best hypothesis, it tends to find a hypothesis close to the best hypothesis in quality.

\subsection{Evaluation on WMT'21 En-De and En-Ru}
\label{sec:noneng}
\begin{figure*}
     \centering
     \begin{subfigure}[b]{0.45\textwidth}
         \centering
         \includegraphics[width=\textwidth]{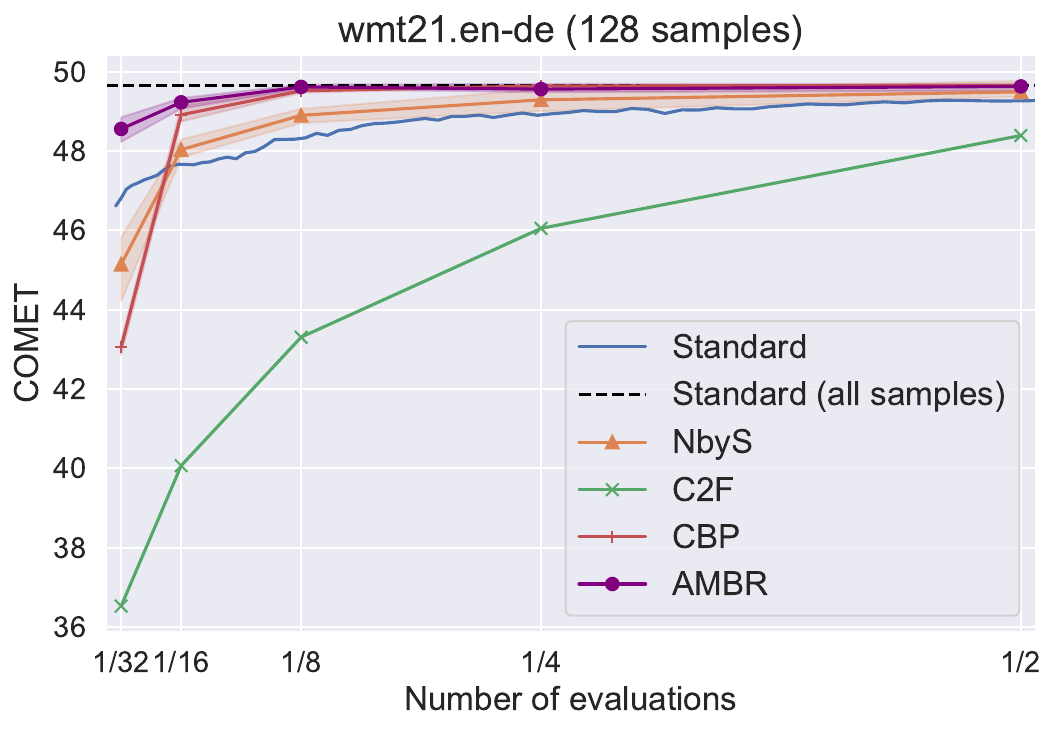}
         \caption{COMET-20 $\uparrow$ (En-De, $N=128$)}
         \label{fig:ende}
     \end{subfigure}
     \hfill
     \begin{subfigure}[b]{0.45\textwidth}
         \centering
         \includegraphics[width=\textwidth]{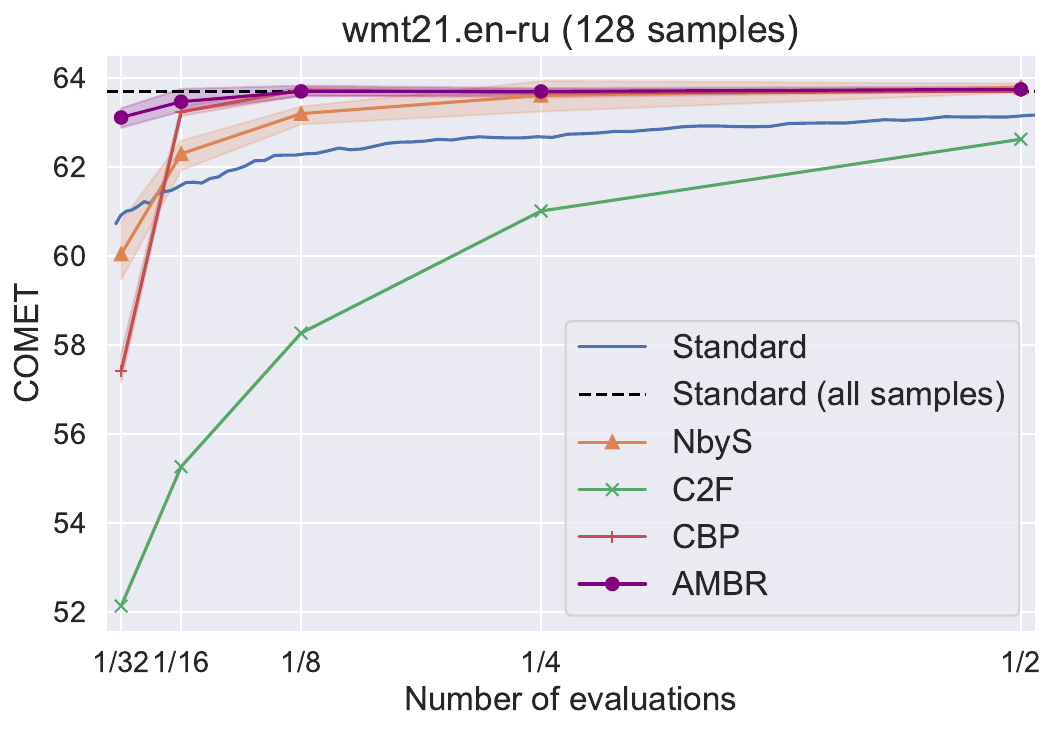}
         \caption{COMET-20 $\uparrow$ (En-Ru, $N=128$)}
         \label{fig:enru}
     \end{subfigure}
    \caption{COMET-20 score and error rate on WMT'21 En-De and En-Ru using WMT 21 En-X model (4.7B). The shaded regions show the minimum and the maximum values over five runs. The error rate is the ratio of selecting a hypothesis different from the standard MBR using all 128 samples. The horizontal axis shows the reduction in the number of evaluations compared to the standard MBR with all 128 samples.}
    \label{fig:enx}
\end{figure*}

To evaluate the performance of AMBR in generating non-English languages, we run experiments on WMT'21 En-De and En-Ru datasets.
We use the WMT 21 En-X model for generating the samples \cite{tran-etal-2021-facebook}. 
For CBP, we search over $r_0 \in \{1, 2, 4\}$ and $\alpha \in \{0.8, 0.9, 0.99\}$.
Other experimental details are the same as in Section \ref{sec:nmt}.
Figure \ref{fig:enx} reports the COMET scores. AMBR and CBP significantly reduce the number of evaluations compared to standard MBR with a marginal drop in the COMET score. NbyS and C2F are less efficient than AMBR and CBP. The performance of AMBR is roughly on par with CBP.
The result of the hyperparameter search for CBP is described in Table \ref{tab:cbp-params-en-x}. The best set of hyperparameters is dependent to the size of the budget.

\begin{table*}
    \centering
\adjustbox{max width=0.95\textwidth}{
\begin{tabular}{ll|rr|rr|rr|rr|rr|rr}
\toprule
\multicolumn{2}{c|}{\#Evaluations} & \multicolumn{2}{c|}{Mean} & \multicolumn{2}{c|}{1/32} & \multicolumn{2}{c|}{1/16} & \multicolumn{2}{c|}{1/8} & \multicolumn{2}{c|}{1/4} & \multicolumn{2}{c}{1/2} \\
$r_0$ & $\alpha$ & COMET & Rank & COMET & Rank & COMET & Rank & COMET & Rank & COMET & Rank & COMET & Rank \\
\midrule\midrule
\multicolumn{14}{c}{WMT'21 En-De ($N=128$)} \\
\midrule
1 & 0.80 & 49.12 & 2 & 47.33 & 2 & 49.45 & 1 & 49.57 & 6 & 49.59 & 7 & 49.65 & 8 \\
1 & 0.90 & 48.88 & 4 & 46.44 & 4 & 49.09 & 5 & 49.58 & 4 & 49.65 & 3 & 49.65 & 7 \\
1 & 0.99 & 48.41 & 6 & 45.17 & 6 & 48.45 & 8 & 49.22 & 9 & 49.55 & 10 & 49.65 & 1 \\
2 & 0.80 & 49.01 & 3 & 46.88 & 3 & 49.26 & 2 & 49.58 & 5 & 49.66 & 1 & 49.65 & 6 \\
2 & 0.90 & 48.83 & 5 & 46.29 & 5 & 48.98 & 6 & 49.61 & 3 & 49.62 & 5 & 49.65 & 2 \\
2 & 0.99 & 48.36 & 7 & 44.90 & 7 & 48.27 & 10 & 49.38 & 8 & 49.58 & 8 & 49.65 & 9 \\
4 & 0.80 & 48.23 & 8 & 42.98 & 10 & 49.23 & 3 & 49.63 & 1 & 49.65 & 2 & 49.65 & 6 \\
4 & 0.90 & 48.16 & 9 & 43.06 & 9 & 48.91 & 7 & 49.52 & 7 & 49.64 & 4 & 49.65 & 4 \\
4 & 0.99 & 48.04 & 10 & 43.32 & 8 & 48.42 & 9 & 49.22 & 10 & 49.60 & 6 & 49.65 & 3 \\
\midrule
\multicolumn{2}{l|}{AMBR} & 49.32 & 1 & 48.56 & 1 & 49.23 & 4 & 49.61 & 2 & 49.56 & 9 & 49.63 & 10 \\
\midrule\midrule
\multicolumn{14}{c}{WMT'21 En-Ru ($N=128$)} \\
\midrule
1 & 0.80 & 63.26 & 2 & 61.80 & 2 & 63.48 & 4 & 63.69 & 9 & 63.66 & 9 & 63.71 & 5 \\
1 & 0.90 & 63.07 & 5 & 60.84 & 5 & 63.41 & 6 & 63.73 & 3 & 63.66 & 8 & 63.70 & 7 \\
1 & 0.99 & 62.65 & 6 & 59.35 & 6 & 62.70 & 10 & 63.75 & 1 & 63.76 & 1 & 63.70 & 9 \\
2 & 0.80 & 63.18 & 3 & 61.25 & 3 & 63.50 & 3 & 63.72 & 5 & 63.72 & 2 & 63.71 & 3 \\
2 & 0.90 & 63.10 & 4 & 60.86 & 4 & 63.54 & 1 & 63.70 & 7 & 63.70 & 4 & 63.72 & 2 \\
2 & 0.99 & 62.63 & 7 & 59.33 & 7 & 62.74 & 8 & 63.74 & 2 & 63.61 & 10 & 63.71 & 4 \\
4 & 0.80 & 62.47 & 8 & 57.76 & 8 & 63.51 & 2 & 63.71 & 6 & 63.68 & 6 & 63.71 & 6 \\
4 & 0.90 & 62.35 & 9 & 57.42 & 10 & 63.23 & 7 & 63.73 & 4 & 63.66 & 7 & 63.70 & 8 \\
4 & 0.99 & 62.27 & 10 & 57.66 & 9 & 62.71 & 9 & 63.58 & 10 & 63.72 & 3 & 63.69 & 10 \\
\midrule
\multicolumn{2}{l|}{AMBR} & 63.54 & 1 & 63.11 & 1 & 63.46 & 5 & 63.70 & 8 & 63.69 & 5 & 63.74 & 1 \\
\bottomrule
\end{tabular}
}
    \caption{Evaluation of confidence-based pruning (CBP) with varying hyperparameters on WMT'21 En-De and En-Ru. $r_0$ is the number of references at the first iteration. $\alpha$ is the threshold of the win rate on pruning. The average COMET-20 score over five runs is reported. Rank denotes the rank of the average score over a set of runs of CBP and AMBR. Mean column reports the average COMET score over 1/32, 1/16, 1/8, 1/4, 1/2.}
    \label{tab:cbp-params-en-x}
\end{table*}

\subsection{Evaluation on M2M100 418M Model}
\label{sec:m2m100}
\begin{figure*}
     \centering
     \begin{subfigure}[b]{0.45\textwidth}
         \centering
         \includegraphics[width=\textwidth]{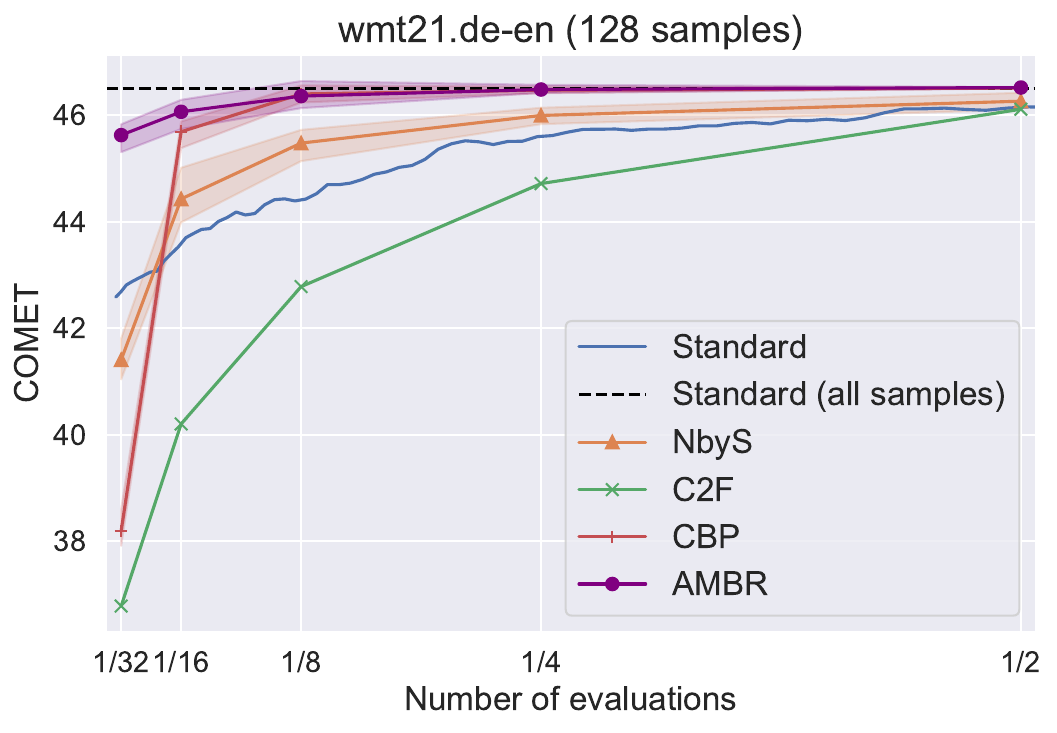}
         \caption{COMET-20 $\uparrow$ (De-En, $N=128$)}
         \label{fig:m2m-deen}
     \end{subfigure}
     \hfill
     \begin{subfigure}[b]{0.45\textwidth}
         \centering
         \includegraphics[width=\textwidth]{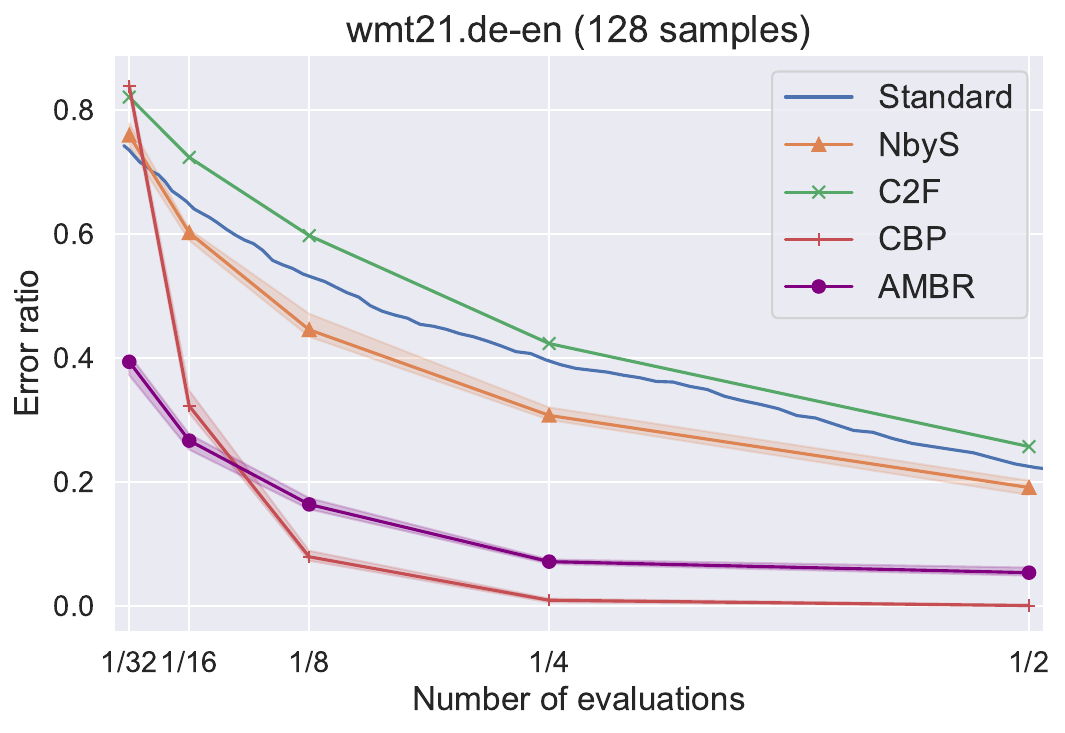}
         \caption{Error Rate $\downarrow$ (De-En, $N=128$)}
         \label{fig:m2m-deen-err}
     \end{subfigure} \\
         \begin{subfigure}[b]{0.45\textwidth}
         \centering
         \includegraphics[width=\textwidth]{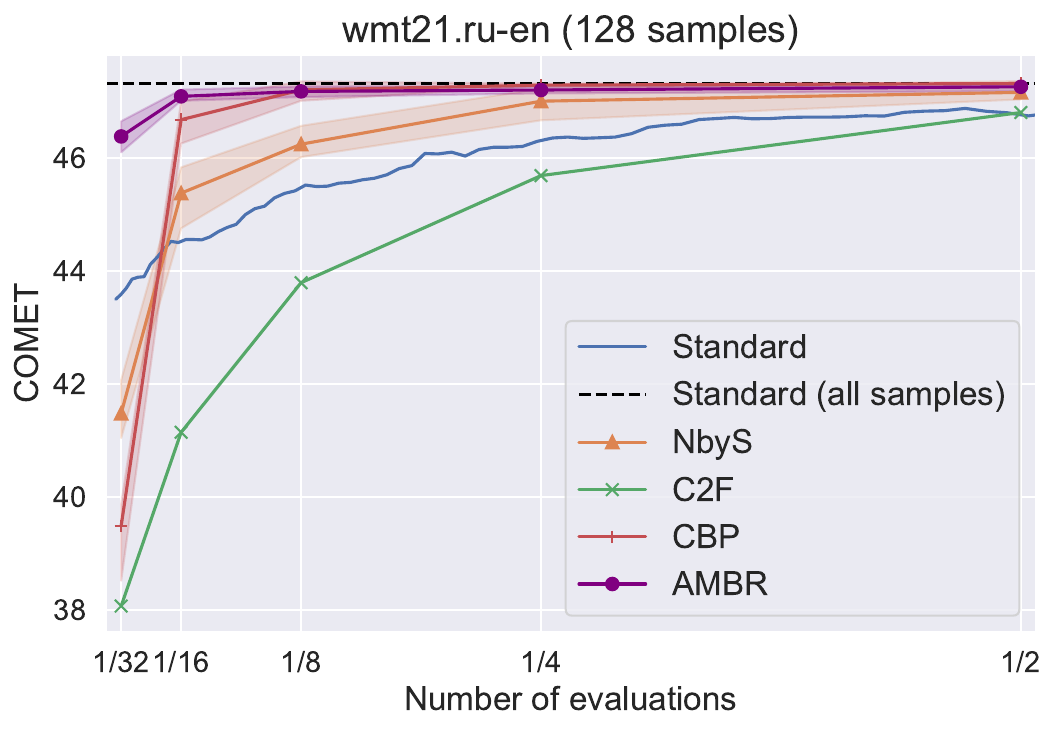}
         \caption{COMET-20 $\uparrow$ (Ru-En, $N=128$)}
         \label{fig:m2m-ruen}
     \end{subfigure}
     \hfill
     \begin{subfigure}[b]{0.45\textwidth}
         \centering
         \includegraphics[width=\textwidth]{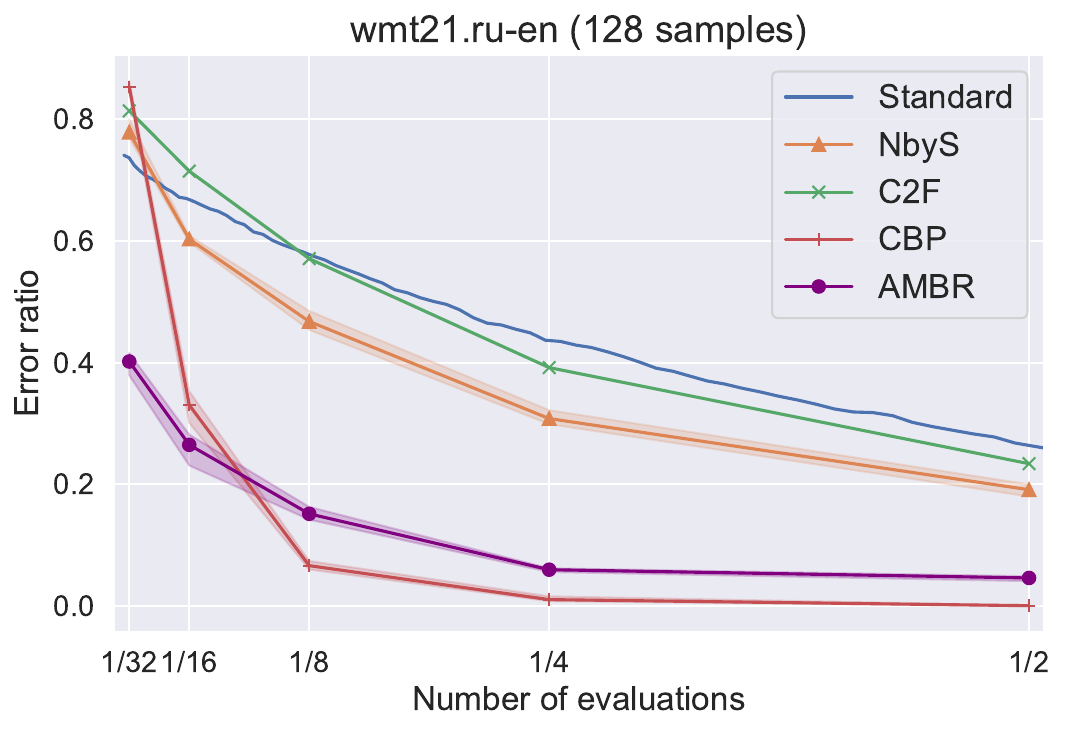}
         \caption{Error Rate $\downarrow$ (Ru-En, $N=128$)}
         \label{fig:m2m-ruen-err}
     \end{subfigure}
    \caption{COMET-20 score and error rate on WMT'21 De-En and Ru-En using the M2M100 418M model. The shaded regions show the minimum and the maximum values over five runs. The error rate is the ratio of selecting a hypothesis different from the standard MBR using all 128 samples. The horizontal axis shows the reduction in the number of evaluations compared to the standard MBR with all 128 samples.}
    \label{fig:m2m100}
\end{figure*}

To compare the performance of the methods on a smaller translation model, we evaluate using the M2M100 418M model.
Figure \ref{fig:m2m100} shows the results. 
Overall, we observe qualitatively the same results as using the WMT 21 En-X model (4.7B). 
AMBR and CBP significantly reduce the number of evaluations compared to standard MBR with a marginal drop in the COMET score. NbyS and C2F are less efficient than AMBR and CBP in WMT'21 tasks. The performance of AMBR is on par with CBP with hyperparameters set by Oracle.

\subsection{Scaling with the Number of Samples on Ru-En}
\label{sec:samples-ruen}

\begin{figure*}
     \centering
     \begin{subfigure}[b]{0.32\textwidth}
         \centering
         \includegraphics[width=\textwidth]{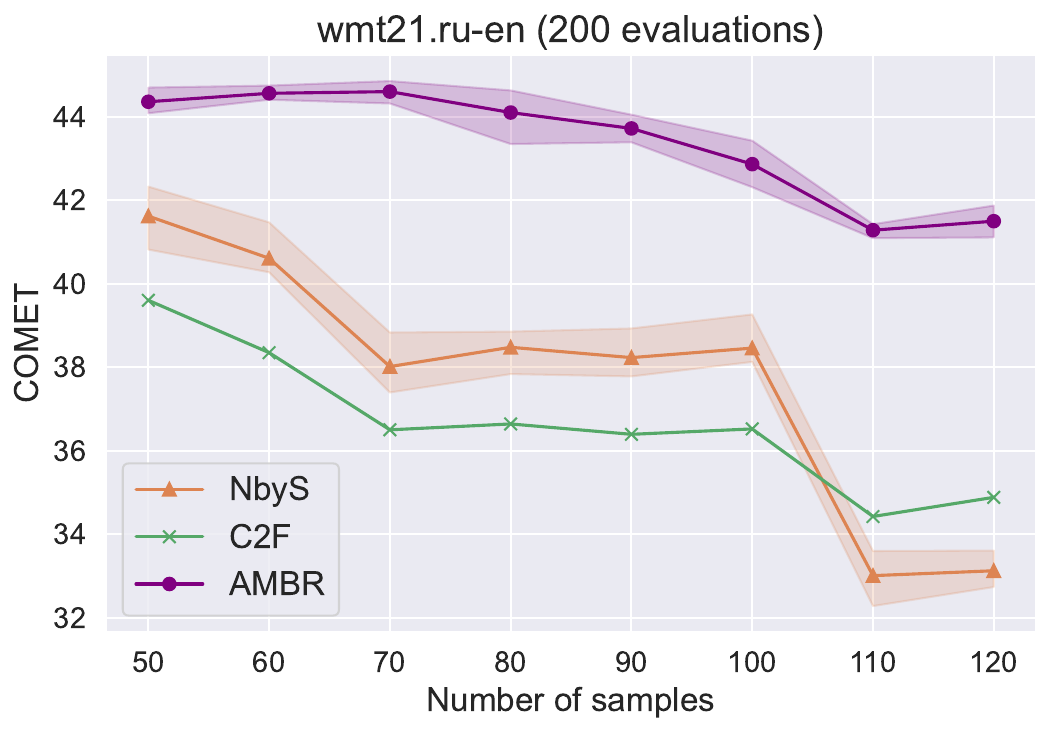}
         \caption{COMET-20 $\uparrow$ (Ru-En, $200$ evaluations)}
         \label{fig:ruen200}
     \end{subfigure}
     \hfill
     \begin{subfigure}[b]{0.32\textwidth}
         \centering
         \includegraphics[width=\textwidth]{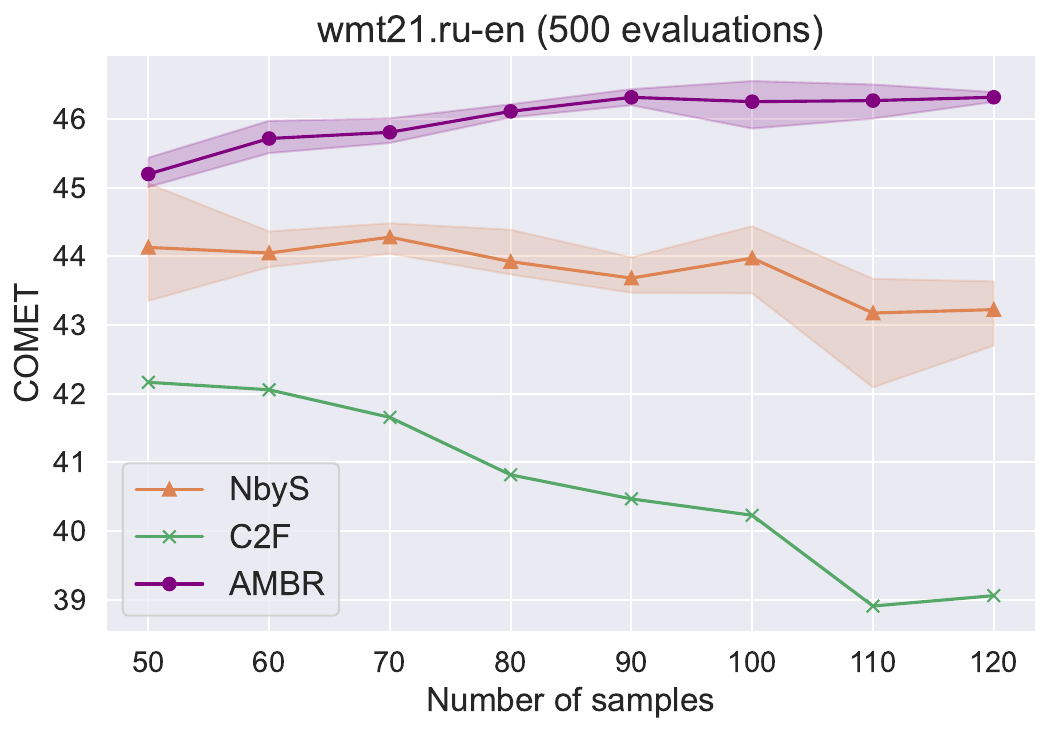}
         \caption{COMET-20 $\uparrow$ (Ru-En, $500$ evaluations)}
         \label{fig:ruen500}
     \end{subfigure}
     \hfill
     \begin{subfigure}[b]{0.32\textwidth}
         \centering
         \includegraphics[width=\textwidth]{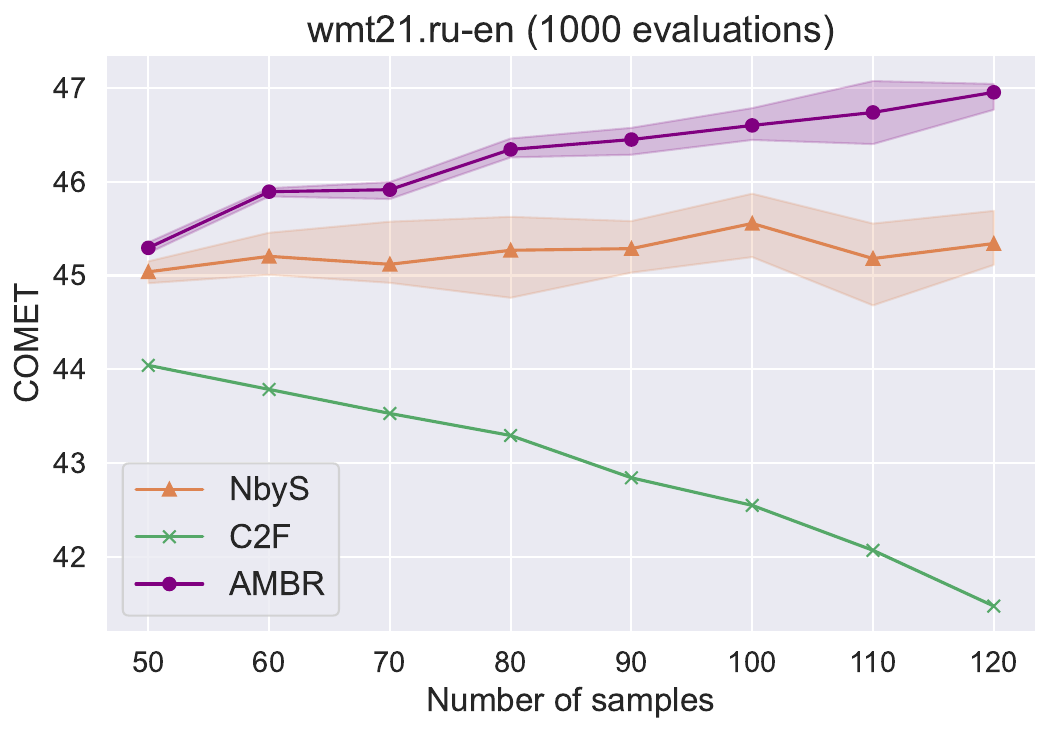}
         \caption{COMET-20 $\uparrow$ (Ru-En, $1000$ evaluations)}
         \label{fig:ruen1000}
     \end{subfigure}
     
     \begin{subfigure}[c]{0.32\textwidth}
         \centering
         \includegraphics[width=\textwidth]{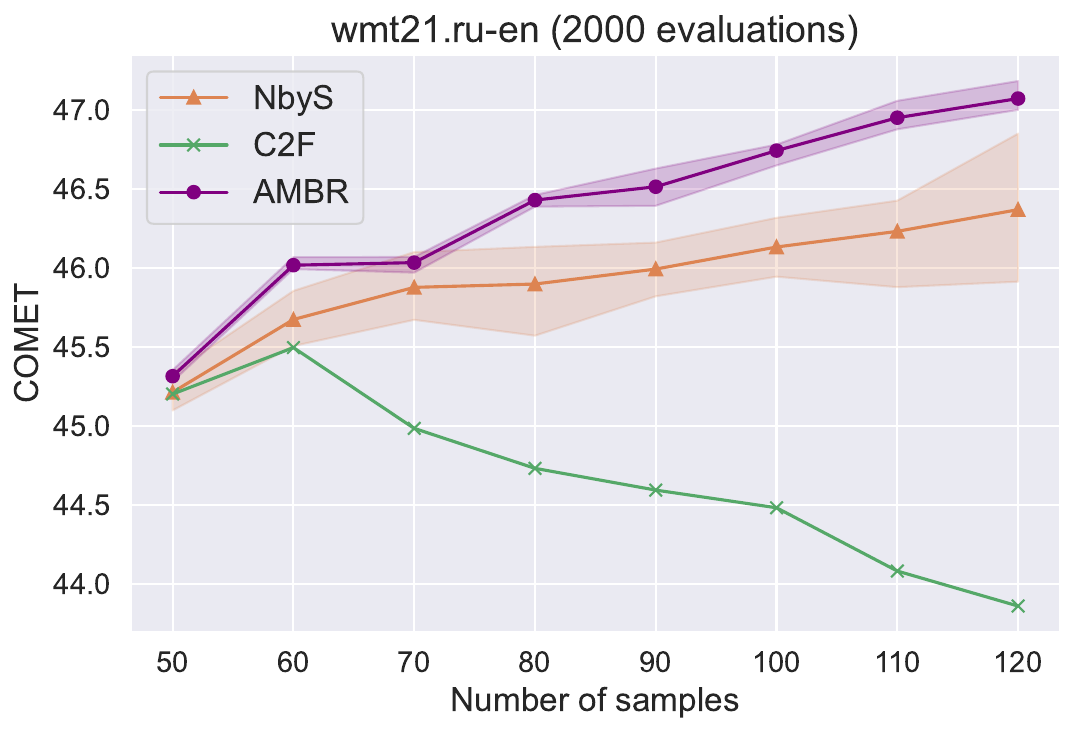}
         \caption{COMET-20 $\uparrow$ (Ru-En, $2000$ evaluations)}
         \label{fig:ruen2000}
     \end{subfigure}
     \hspace{10pt}
     \begin{subfigure}[c]{0.32\textwidth}
         \centering
         \includegraphics[width=\textwidth]{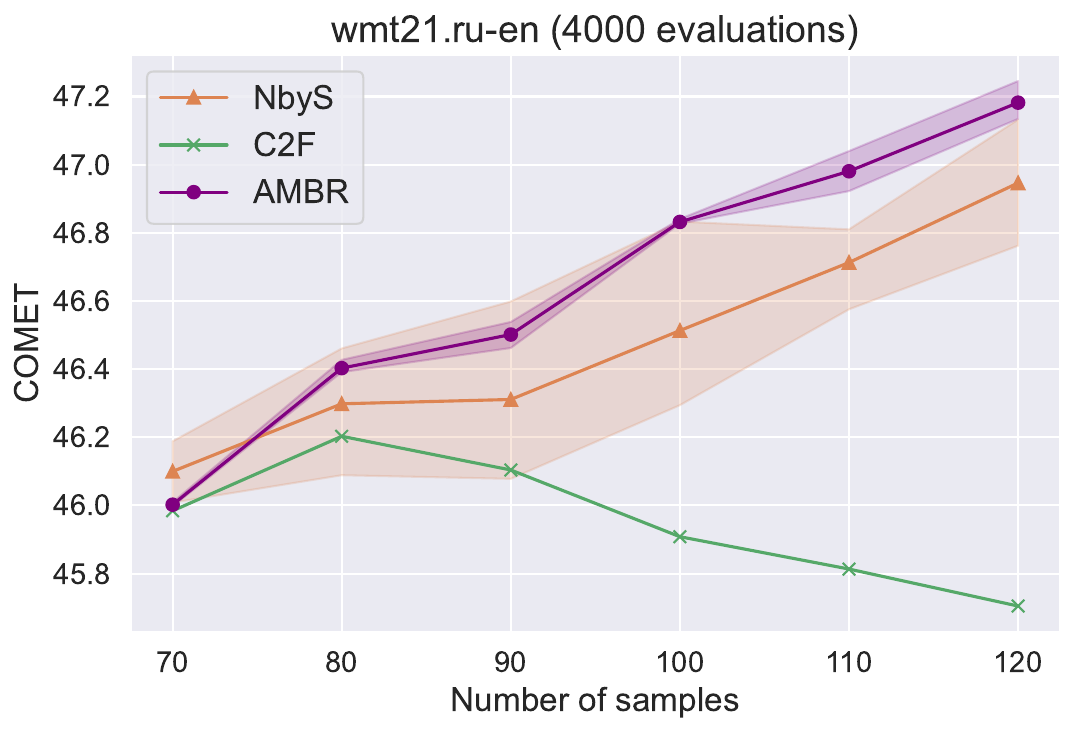}
         \caption{COMET-20 $\uparrow$ (Ru-En, $4000$ evaluations)}
         \label{fig:ruen4000}
     \end{subfigure}
    \caption{COMET-20 score on WMT'21 Ru-En with varying number of samples using M2M100 418M model. The shaded regions show the minimum and the maximum values over five runs.}
    \label{fig:samples-ruen}
\end{figure*}

Figure \ref{fig:samples-ruen} shows the result on WMT'21 Ru-En with varying sample sizes with a fixed evaluation budget on the M2M100 418M model. We observe the same trends as in WMT'21 De-En (Figure \ref{fig:samples}). AMBR scales with the number of samples if there is enough evaluation budget. 

\section{Pretrained Models used in the Experiments}

We list the pretrained models we used in the experiments in Table \ref{tab:models}.

\begin{table*}
    \centering
    \adjustbox{max width=\textwidth}{
    \begin{tabular}{c|l}
    \toprule
        WMT'21 (Section \ref{sec:nmt}) & \citet{tran-etal-2021-facebook} \url{https://huggingface.co/facebook/wmt21-dense-24-wide-x-en} \\
        WMT'21 (Section \ref{sec:nmt}) & \citet{10.5555/3546258.3546365} \url{https://huggingface.co/facebook/m2m100_418M} \\
        WMT'21 (Section \ref{sec:noneng}) & \citet{tran-etal-2021-facebook} \url{https://huggingface.co/facebook/wmt21-dense-24-wide-en-x} \\
        SAMSum (Section \ref{sec:summarization}) & \url{https://huggingface.co/philschmid/bart-large-cnn-samsum} \\
        XSum (Section \ref{sec:summarization}) & \citet{lewis-etal-2020-bart} \url{https://huggingface.co/facebook/bart-large-xsum} \\
        MS COCO (Section \ref{sec:captioning}) & \citet{pmlr-v202-li23q} \url{https://huggingface.co/Salesforce/blip2-flan-t5-xl-coco} \\
        MS COCO (Section \ref{sec:captioning}) & (CLIPScore) \citet{hessel-etal-2021-clipscore} \url{https://huggingface.co/openai/clip-vit-large-patch1} \\
    \bottomrule
    \end{tabular}
    }
    \caption{List of pretrained models we used in the experiments.}
    \label{tab:models}
\end{table*}

\end{document}